\documentclass{article}

\usepackage[preprint]{ms}

\usepackage{natbib} % has a nice set of citation styles and commands
\usepackage[utf8]{inputenc} % allow utf-8 input
\usepackage[T1]{fontenc}    % use 8-bit T1 fonts
\usepackage{hyperref}       % hyperlinks
\usepackage{url}            % simple URL typesetting
\usepackage{booktabs}       % professional-quality tables
\usepackage{amsfonts}       % blackboard math symbols
\usepackage{nicefrac}       % compact symbols for 1/2, etc.
\usepackage{microtype}      % microtypography
\usepackage{xcolor}         % colors

\usepackage{amsmath}
\usepackage{amssymb}
\usepackage{bbm}
\usepackage{mathtools}
\usepackage{amsthm}
\usepackage{amsfonts}
\usepackage{centernot}
\usepackage{xspace}

\newcommand{\ica}{\textsc{Ica}\xspace}
\newcommand{\pcl}{\textsc{Pcl}\xspace}
\newcommand{\tcl}{\textsc{Tcl}\xspace}
\newcommand{\kvae}{\textsc{KalmanVae}\xspace}
\newcommand{\betavae}{$\beta$-\textsc{Vae}\xspace}
\newcommand{\slowvae}{\textsc{SlowVae}\xspace}
\newcommand{\ivae}{i\textsc{Vae}\xspace}
\newcommand{\leap}{\textsc{Leap}\xspace}

\newcommand{\leaplin}{\textsc{Leap-Lin}\xspace}
\newcommand{\leapnp}{\textsc{Leap-Np}\xspace}
\newcommand{\leapnpmlp}{\textsc{Leap-Np-Mlp}\xspace}
\newcommand{\mlpdynone}{$1$\textsc{-Mlp-Dyn}\xspace}
\newcommand{\mlpdynK}{$K$\textsc{-Mlp-Dyn}\xspace}

\renewcommand{\vec}[1]{\mathbf{#1}}
\newcommand*{\vd}{\vec{d}}
\newcommand*{\ve}{\vec{e}}
\newcommand*{\vf}{\vec{f}}
\newcommand*{\vr}{\vec{r}}
\newcommand*{\vs}{\vec{s}}
\newcommand*{\vu}{\vec{u}}
\newcommand*{\vx}{\vec{x}}
\newcommand*{\vz}{\vec{z}}
\newcommand{\N}{\mathcal{N}}
\newcommand{\pth}{p_{\theta}}
\newcommand{\qph}{q_{\phi}}
\newcommand{\D}{\mathop{}\!\mathrm{d}}
\newcommand{\id}{\mathrm{id}}
\newcommand{\x}{\mathbf{x}}
\newcommand{\br}{\mathbf{r}}
\newcommand{\z}{\mathbf{z}}
\newcommand{\s}{\mathbf{s}}
\newcommand{\f}{\mathbf{f}}

\newcommand{\w}{\mathbf{w}}
\newcommand{\bk}{{\mathbf{k}}}

\newcommand{\faug}{{\f_{\texttt{aug}}}}

\newcommand{\gaug}{{\g_{\texttt{aug}}}}

\newcommand{\kaug}{{\bk_{\texttt{aug}}}}
\newcommand{\g}{{\mathbf{g}}}
\newcommand{\bv}{\mathbf{v}}

\newcommand{\hatz}{\hat{\z}}

\newcommand{\hatf}{{\hat{\f}}}
\newcommand{\hatg}{{\hat{\g}}}
\newcommand{\hats}{\hat{\s}}

\newcommand{\hatfaug}{{\hat{\mathbf{f}}_{\texttt{aug}}}}

\newcommand{\hatgaug}{{\hat{\g}_{\texttt{aug}}}}
\newcommand{\h}{{\mathbf{h}}}
\newcommand{\bH}{{\mathbf{H}}}
\newcommand{\K}{{\mathbf{K}}}

\newcommand{\bu}{\mathbf{u}}

\newcommand{\R}{\mathbb{R}}
\newcommand{\indep}{\perp \!\!\! \perp}

\newcommand{\logs}[1]{q_{#1}(s_{#1 t}, \bu)}

\newcommand{\logz}[1]{\eta_{#1}(z_{#1 t}, \bu)}

\newcommand{\hatlogs}[1]{\hat{q}_{#1}(\hat{s}_{#1 t}, \bu)}
\newcommand{\hatlogz}[1]{\hat{\eta}_{#1}(\hat{z}_{#1 t}, \bu)}

\DeclareMathOperator{\EX}{\mathbb{E}}

\newcommand{\zpair}{
\begin{bmatrix}
    \z_t \\ \z_{t-1} 
\end{bmatrix}
}
\newcommand{\xpair}{
\begin{bmatrix}
    \x_t \\ \x_{t-1} 
\end{bmatrix}
}
\newcommand{\fauginpair}{
\begin{bmatrix}
    \s_t \\ \z_{t-1} 
\end{bmatrix}
}
\newcommand{\faugimpl}{
\begin{bmatrix}
    \f(\z_{t-1}, \s_{t}) \\ \z_{t-1} 
\end{bmatrix}
}

\newcommand{\augvec}[2]{
\begin{bmatrix}
    #1 \\ #2
\end{bmatrix}
}

\newcommand{\Ttrain}{\ensuremath{T_{\text{train}}}}
\newcommand{\Tdyn}{\ensuremath{T_{\text{dyn}}}}

\newcommand{\Tfuture}{\ensuremath{T_{\text{future}}}}
\newcommand{\ztrain}{\ensuremath{\bar{\z}_{\text{train}}}}
\newcommand{\strain}{\ensuremath{\bar{\s}_{\text{train}}}}
\newcommand{\xfuture}{\ensuremath{\bar{\x}_{\text{future}}}}
\newcommand{\zfuture}{\ensuremath{\bar{\z}_{\text{future}}}}
\newcommand{\xfutureh}[1]{\ensuremath{\bar{\x}_{\text{future}[#1]}}}
\newcommand{\zfutureh}[1]{\ensuremath{\bar{\z}_{\text{future}[#1]}}}
\newcommand{\mcc}{\textsc{Mcc}}
\newcommand{\mse}{\textsc{Mse}}
\newcommand{\mccz}{\ensuremath{\textsc{Mcc}[\ztrain]}}
\newcommand{\mccs}{\ensuremath{\textsc{Mcc}[\strain]}}
\newcommand{\mcczfuture}{\ensuremath{\textsc{Mcc}[\zfuture]}}
\newcommand{\mcczfutureh}[1]{\ensuremath{\textsc{Mcc}[\zfutureh{#1}]}}
\newcommand{\msexfuture}{\ensuremath{\textsc{Mse}[\xfuture]}}
\newcommand{\msexfutureh}[1]{\ensuremath{\textsc{Mse}[\xfutureh{#1}]}}

\newcommand{\elbo}{\mathcal{L}}
\newcommand{\elboR}{\elbo_\text{R}}
\newcommand{\elboKL}{\elbo_\text{KL}}

\newcommand{\smallertext}[1]{\scalebox{0.7}{#1}}
\newcommand{\std}[1]{\smallertext{($\pm #1$)}}

\usepackage[capitalize]{cleveref}

\usepackage{tikz}
\usetikzlibrary{arrows.meta, positioning, shapes, shapes.geometric, fit, calc, decorations.pathreplacing}
\usepackage{xcolor}

\usepackage{caption}
\usepackage{subcaption}
\usepackage{bbm}
\usepackage{multirow}
\usepackage{enumitem}
\usepackage{wrapfig}

\usepackage{algorithm}
\usepackage{algorithmic}

\title{Identifying latent state transition in non-linear dynamical systems}

\author{%
    \c{C}a\u{g}lar H{\i}zl{\i} \\
    Aalto University\\
    \texttt{caglar.hizli@aalto.fi} \\
    \And
    \c{C}a\u{g}atay Y{\i}ld{\i}z \\
    University of Tübingen \\
    Tübingen AI Center \\
    \And
    Matthias Bethge \\
    University of Tübingen \\
    Tübingen AI Center \\
    \And
    ST John \\
    Aalto University\\
    \And
    Pekka Marttinen \\
    Aalto University\\
}

\begin{document}

	\maketitle

	\begin{abstract}
		This work aims to improve generalization and interpretability of dynamical systems by recovering the underlying lower-dimensional latent states and their time evolutions.		
		Previous work on disentangled representation learning within the realm of dynamical systems focused on the latent states, possibly with linear transition approximations. As such, they cannot identify nonlinear transition dynamics, and hence fail to reliably predict complex future behavior.
		Inspired by the advances in nonlinear \ica, we propose a state-space modeling framework in which we can identify not just the latent states but also the unknown transition function that maps the past states to the present.
		We introduce a practical algorithm based on variational auto-encoders and empirically demonstrate in realistic synthetic settings that we can (i) recover latent state dynamics with high accuracy, (ii) correspondingly achieve high future prediction accuracy, and (iii) adapt fast to new environments.\looseness-1
	\end{abstract}

	\section{Introduction}
\begin{figure}
		\centering
		\includegraphics[width=1\linewidth]{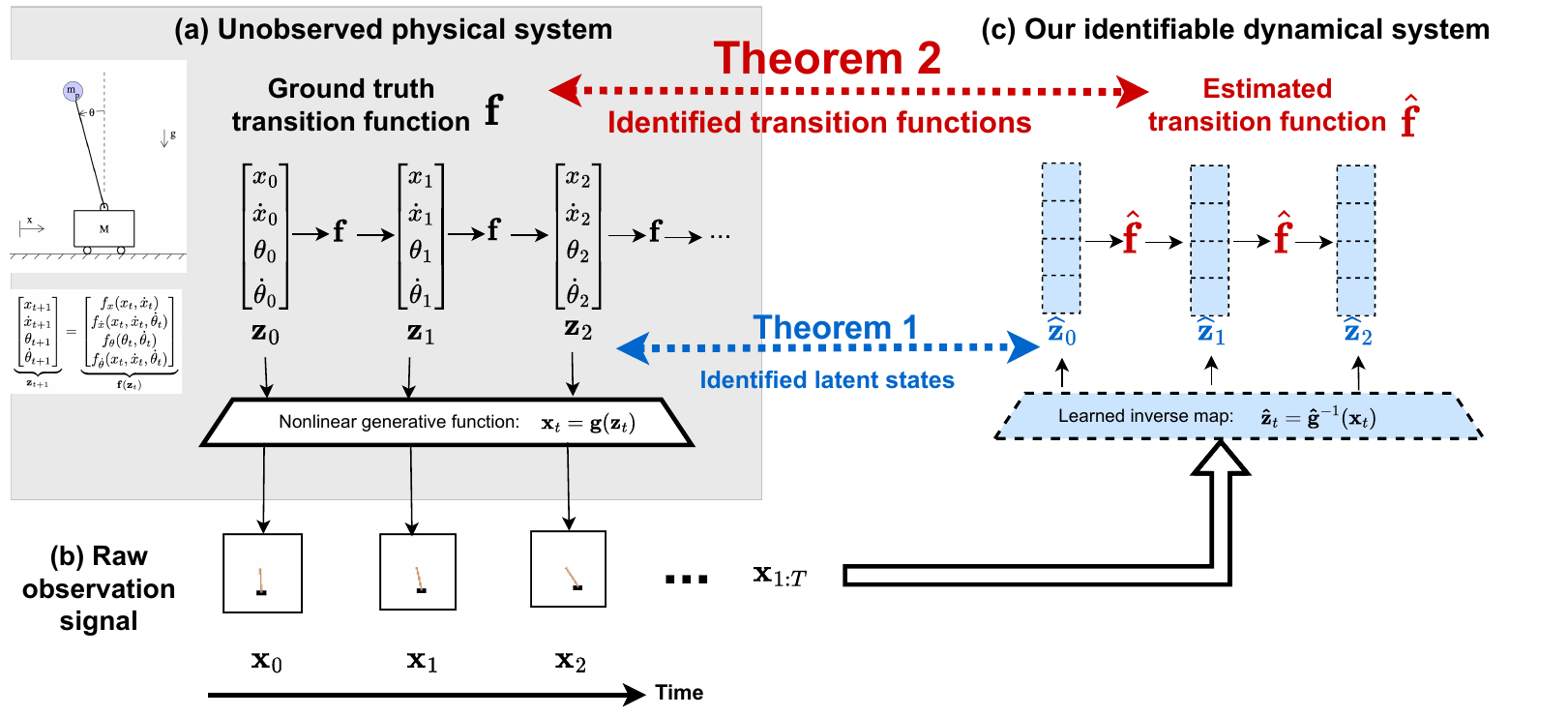}
		\caption{Sketch of our method and main theoretical contribution. \textbf{(a)} We assume an underlying \textit{unobserved} dynamical system, e.g., a cartpole, where the full state is composed of the cart position and velocity, and the angle and angular velocity of the pole: $[x, \dot{x}, \theta, \dot{\theta}]$. \textbf{(b)} We partially observe the system as a sequence of video frames, which are used as input to our method. \textbf{(c)} We learn an inverse generative function that maps the raw observation signals to the latent state variables, as well as a transition function that maps the past latent states to the present latent state. {\color{blue} Identifiability of the latent states} is ensured by {\color{blue} Theorem 1} \citep{yao2021learning}. In addition to this, our main contribution is {\color{red} the identifiability of the transition function} ensured by {\color{red} Theorem 2}.}
		\label{fig:fig1}
	\end{figure}

	We focus on the problem of understanding the underlying states of a target dynamical system from its low-level, high-dimensional sensory measurements. This task is prevalent across various fields, including reinforcement learning \citep{hafner2019dream} and robotics \citep{levine2016end}. As a running example of such a system, we consider a drone controlled by an autonomous system. Here, the observational data would be a video stream (\cref{fig:fig1}~\textbf{(b)}) instead of the full state of the system comprised of absolute position, velocity, and acceleration in $3D$ (\cref{fig:fig1} \textbf{(a)}). This system may have additional variables influencing the state evolution, e.g., the strength and direction of the wind at any time or drone motor torque set by the controller. The main objective of this work is to learn latent representations and state transition functions that would be useful for downstream tasks, e.g., computing the control signals for the optimal transport of the drone from point A to point B.
	
	\paragraph{Identifiability.} Due to the partially observed nature of these problems, learning dynamics directly in the data space (e.g., pixel space) is not feasible, and previous works often focus on learning \emph{latent dynamical systems} \citep{hafner2019learning}. However, such latent models commonly are not guaranteed to recover the true underlying states and transitions (\emph{non- identifiability}), which results in entangled representations, lack of generalization across new domains, and poor interpretability \citep{schmidhuber1992learning,bengio2013representation}.
	Identifiable representation learning aims to address these challenges by learning the underlying factors of variation in the true generative model \citep{hyvarinen1999nonlinear}. Most existing methods \citep{hyvarinen2016unsupervised, hyvarinen2017nonlinear, hyvarinen2019nonlinear,khemakhem2020variational,klindt2020towards} assume mutually independent components that do not affect each other. This is unrealistic for dynamical systems as the present state of the system depends on the past states, i.e., the transition function propagates the system state by (nonlinearly) mixing the past state components \citep{morioka2021independent, yao2021learning, yao2022temporally}.
	
	\paragraph{Identifiability in dynamical systems.}
	Recently, \citet{yao2021learning,yao2022temporally} showed that under certain assumptions, it is possible to \emph{identify} or recover the true unobserved latent states in a dynamical system (up to component-wise transformations). 
	\citet{klindt2020towards} show that temporally sparse transitions could recover the true latent variables up to trivial transformations. \citet{morioka2021independent} introduce a framework to estimate the `innovations' or process noise, which represents the stochastic impulses fed to a dynamical process. 

 \citet{yao2021learning} recover latent states and identify their temporal relations from sequential data by utilizing non-stationarity noise. Later, \citet{yao2022temporally} extend their framework to instead exploit the autocorrelation between the past and the present latent states, while including factors modulating the dynamics and generative functions.
 While these attempts lead to provably identifiable representations, they only propose non-parametric or linear approximations to the unobserved state transition function. While their nonparametric approximations cannot be unrolled over time, a linear model falls short of predicting future states of complicated systems (as we demonstrate empirically in our experiments).
	
	\paragraph{Our contributions.}
	We present the first framework that allows for the \textit{identification of the unknown transition function} alongside latent states and the generative function (see \cref{fig:fig1}).
	Following the previous works \citep{klindt2020towards, yao2021learning, yao2022temporally}, we first establish the identifiability of the latent states (\cref{fig:fig1}: {\color{blue} Theorem 1}). 
	Different from these works, our framework allows estimating the process noise, representing the random impulses fed to a dynamical system. Inspired by \citet{morioka2021independent}, we show that the estimation of the correct transition function is ensured by restricting the process noise and the transition function (\cref{fig:fig1}: {\color{red} Theorem 2}). These restrictions form our main assumptions for identifiability: (i) the process noise shows non-stationarity or autocorrelation driven by an auxiliary variable and (ii) a trivially augmented transition function is bijective.
	
	In our formulation, the process noise is encoded from observations similar to \citet{franceschi2020stochastic}.
	While previous works perform backward prediction of noise from present and past states \citep{yao2021learning,yao2022temporally}, we show that encoding the process noise and predicting future latent states from the past latent states and process noises is the key to achieving an identifiable transition function.
	We propose a straightforward evidence lower bound that allows us to recover true underlying factors in the limit of infinite data. 
	Our empirical findings show that our framework manages to predict the future states of an unknown system and is easier to adapt to novel environments.
	
	\section{An identifiable dynamical system framework}
	\label{sec:theory}
	This section starts with the notation and our generative dynamical model that leads to identifiable variables and functions under certain assumptions. Next, we state the assumptions to achieve identifiability, followed by our main theoretical contribution.
	
	\subsection{Our generative model}
	\label{ssec:gen-model}
	We are interested in inferring latent dynamical systems from high-dimensional sensory observations $\x_{1:T}$, where $t$ is the time index and $\x_t \in \R^D$. As usual, we assume a sequence of latent states $\z_{1:T}$, with $\z_t \in \R^K$, are instantaneously mapped to observations via a generative function $\g: \R^K \rightarrow \R^D$:
	$$ \x_t = \g(\z_t).$$
	Without loss of generality, the latent states $\z_{1:T}$ evolve according to Markovian dynamics:
	$$ \z_t = \f(\z_{t-1},\s_{t}), $$
	where $\f: \R^{2K} \rightarrow \R^K$ is an auto-regressive transition function and $\s_t \in \R^K$ corresponds to additional variables influencing the dynamics, e.g., random forces acting on the system or control signals.
	
	\paragraph{Augmented dynamics}
	For the ease of notation in the rest of the paper, we introduce the following generative process with augmented transition and generative functions \citep{morioka2021independent}:
	\begin{align}
		\label{eq:gen-ic}
		\z_0 &\sim p_{\z_0}(\z_0), &&\texttt{\# initial state} \\
		\label{eq:gen-noise}
		\s_t &\sim p_{\s | \bu}(\s_t | \bu) = \prod_{k} p_{s_{k} | \bu}(s_{kt} | \bu), \quad \forall t \in 1, \ldots, T, &&\texttt{\# process noise} \\
		\label{eq:gen-transition}
		\zpair &= \faug \left( \fauginpair \right) = \faugimpl, \quad \forall t \in 1, \ldots, T, &&\texttt{\# state transition} \\
		\label{eq:gen-obs}
		\xpair &= \gaug \left( \zpair \right) = \augvec{\g(\z_t)}{\g(\z_{t-1})}, \qquad \forall t \in 2, \ldots, T. &&\texttt{\# observation mapping} 
	\end{align}
	Above, $\bu$ is an auxiliary variable modulating the noise distribution $p_{\s|\bu}$. Recent nonlinear independent component analysis (\textsc{Ica}) works have achieved strong identifiability results by exploiting such auxiliary variables to obtain non-stationarities \citep{hyvarinen2016unsupervised} and autocorrelations \citep{hyvarinen2017nonlinear}.
	Inspired by these works, we consider the following cases:
	\begin{itemize}
		\item Setting $\bu$ to an observed regime index leads to a \textbf{nonstationary process noise}. A real-world example would be a flying drone under different precipitation conditions, which can be observed up to a noise level.
		\item Setting $\bu=\s_{t-1}$ implies an \textbf{autocorrelated noise process}. A real-world example would be a flying drone in a windy environment where the wind speed or direction changes continuously.
	\end{itemize}

	\subsection{Identifiability theory} 
	\label{ssec:idf_theory} 
	Let $\mathcal{M} = (\faug, \gaug, p_{\s|\bu})$ denote the ground-truth model. We learn a model $\hat{\mathcal{M}} = (\hatfaug, \hatgaug, \hat{p}_{\s|\bu})$ by fitting the observed sequences. We make the following assumptions:
	
	\begin{enumerate}
		\item [\textit{(A0)}] \textbf{Distribution matching \citep{klindt2020towards, yao2021learning, yao2022temporally}} The learned and the ground-truth observation densities match everywhere:
		\begin{align}
			p_{\faug, \gaug, p_{\s|\bu}}(\{\x_t\}_{t=1}^T) = p_{\hatfaug, \hatgaug, \hat{p}_{\s|\bu}}(\{\x_t\}_{t=1}^T), \forall \x_t \in \mathcal{X}.
		\end{align}
        This is a standard assumption in the nonlinear \ica literature. It ensures that the model $\hat{\mathcal{M}}$ is sufficiently flexible, e.g., a neural network, that it learns the correct distribution in the limit of infinite data.
		\item [\textit{(A1)}] \textbf{Injectivity and bijectivity \citep{morioka2021independent}} The generator functions $\g$ and $\hatg$ are injective, which implies that the augmented generative functions $\gaug, \hatgaug$ are injective. The augmented dynamics functions $\faug, \hatfaug$ are bijective. 
  
          In contrast to \citet{morioka2021independent}, which use an augmented transition model on \textit{observations} $(\x_{t-1}, \x_t)$, our formulation captures the functional dependence between a \textit{latent pair} $(\z_{t-1}, \z_t)$ and the process noise $\s_t$.
		\item [(\textit{A2})] \textbf{Decomposed transitions \citep{klindt2020towards, yao2021learning, yao2022temporally, song2023temporally}} Each dimension of the transition function $\{f_k\}_{k=1}^K$ is influenced by a single process noise variable $s_{kt}$. The output is a single latent variable $z_{kt}$:
        \begin{align}
            z_{kt} &= f_k (\z_{t-1}, s_{kt}), \quad \text{for } k \in 1, \ldots, K \text{ and } t \in 1, \ldots, T.
        \end{align}
		\item [\textit{(A3)}] \textbf{Conditionally independent noise.} Let $\logs{k} = \log p(s_{kt} | \bu)$ denote the conditional density of the noise variable $s_{kt}$. Let $\logz{k} = \log p(z_{kt} | \z_{t-1}, \bu)$ denote the conditional density of the state variable $z_{kt}$. Conditioned on the auxiliary variable $\bu$, we assume:
		\begin{itemize}
			\item Each noise variable $\s_t \in \R^K$ is independent over its dimensions $s_{1t}, \ldots, s_{Kt}$: 
			\begin{equation}
				\log p(\s_t | \bu) = \sum_{k=1}^K \underbrace{\log p(s_{kt} | \bu)}_{\logs{k}} = \sum_{k=1}^K \logs{k}, \quad \forall t \in 1, \ldots, T,
			\end{equation}
			\item the past latent state $\z_{t-1}$ and the present noise $\s_t$ are independent: 
			$\s_t \indep \z_{t-1} | \bu$.
		\end{itemize}
		Since $\s_t \indep \z_{t-1} | \bu$ and each dimension of the transition function $f_k(\z_{t-1}, s_{kt})$ is a function of only a single (conditionally) independent noise variable $s_{kt}$, the conditional density of the latent pair $(\z_t,\z_{t-1})$ also factorizes:
		\begin{align}
			\log p(\z_t | \z_{t-1}, \bu) = \sum_{k=1}^K \underbrace{\log p(z_{kt} | \z_{t-1}, \bu)}_{\logz{k}} = \sum_{k=1}^K \logz{k}.
		\end{align}
		The same is assumed for the learned conditional density: $\log p(\hatz_t | \hatz_{t-1}, \bu) = \sum_{k=1}^K \log p(\hat{z}_{kt} | \hatz_{t-1}, \bu).$ 
  
        This assumption generalizes the nonstationary noise assumption in \citet{yao2021learning} which consider the more specific case of setting the auxiliary variable $\bu$ to the regime index.
		\item [\textit{(A4)}] \textbf{Sufficient variability of latent state $\z_t$.} For any $\z_t$, there exist some $2K$ values of $\bu$: $\bu_1, \ldots, \bu_{2K}$, such that the $2K$ vectors
		\begin{equation}
			\bv(\z_t, \bu_1), \ldots, \bv(\z_t, \bu_{2K})
		\end{equation}
		are linearly independent for some index $l$ of the auxiliary variable $\bu$, where
		\begin{align}
			\bv(\z_t, \bu) = \left( \frac{\partial^2 \eta_{1}(z_{1t}, \bu)}{\partial z_{1t} \partial u_{l}}, \cdots, \frac{\partial^2 \eta_{K}(z_{Kt}, \bu)}{\partial z_{Kt} \partial u_{l}}, \frac{\partial^3 \eta_{1}(z_{1t}, \bu)}{\partial z^2_{1t} \partial u_{l}}, \cdots, \frac{\partial^3 \eta_{K}(z_{Kt}, \bu)}{\partial z^2_{Kt} \partial u_{l}} \right) \in \R^{2K}.
		\end{align}
        This assumption is a generalisation of the sufficient variability assumption in \citet{yao2021learning} where the conditional density $\logz{k}$ conditions on a specific regime index rather than a general auxiliary variable.
        It implies that the dependence of the latent states $\z_t$ on the auxiliary variable $\bu$ is not too 'simple', meaning that the corresponding conditional log densities $\logz{k}$ change with respect to $\bu$ sufficiently differently for each value of $\bu_1, \ldots, \bu_{2K}$, e.g., eliminating 'simple' rotation-invariant Gaussian densities \citep{hyvarinen2019nonlinear,hyvarinen2023nonlinear}.
		\item [\textit{(A5)}] \textbf{Sufficient variability of process noise $\s_t$.} For any $\s_t$, there exist some $2K$ values of $\bu$: $\bu_1, \ldots, \bu_{2K}$, such that the $2K$ vectors
		\begin{equation}
			\w(\s_t, \bu_1), \ldots, \w(\s_t, \bu_{2K})
		\end{equation}
		are linearly independent for some index $l$ of the auxiliary variable $\bu$, where
		\begin{align}
			\w(\s_t, \bu) = \left( \frac{\partial^2 q_{1}(s_{1t}, \bu)}{\partial s_{1t} \partial u_l}, \cdots, \frac{\partial^2 q_{K}(s_{Kt}, \bu)}{\partial s_{Kt} \partial u_l}, \frac{\partial^3 q_{1}(s_{1t}, \bu)}{\partial s^2_{1t} \partial u_l}, \cdots, \frac{\partial^3 q_{K}(s_{Kt}, \bu)}{\partial s^2_{Kt} \partial u_l} \right) \in \R^{2K}.
		\end{align}
        Similar to \textit{(A3)}, this assumption ensures that the noise density contains enough structure in the form of autocorrelation or non-stationarity, where the process noise density changes with respect to $\bu$ sufficiently differently for each value of $\bu_1, \ldots, \bu_{2K}$.
	\end{enumerate}
	
	\paragraph{Remark} \citet{yao2021learning} used nonstationarity of the process noise for identifiability of the conditionally independent latent states (\textbf{Theorem 1}). If the variable $\bu$ is an observed categorical variable (e.g., domain indicator), the assumptions \textit{(A4, A5)} can be written in an alternative form without partial derivatives with respect to $u_l$ \citep{hyvarinen2019nonlinear} (see \cref{app:sec:alternative-assump} for the alternative version).
 
 \subsection{Main theoretical contribution}
	In this section, we state our main theoretical contribution, that is, the identifiability result for the dynamical function $\f$ (\textbf{Theorem 2}). For completeness, we start with a theorem on the identifiability result for the conditionally independent latent states $\z_t | \z_{t-1}, \bu$ (\textbf{Theorem 1}), which \citet{yao2021learning} established for the nonstationary noise case. The proofs are detailed in \cref{app:sec:proof1,app:sec:proof2}.
	
	\paragraph{Theorem 1 (Identifiability of latent states $\z_{1:T}$) \citep{yao2021learning}:} Under assumptions \textit{(A0, A1, A2, A3, A4)}, latent states $\z_t = \h(\hatz_t)$ are identifiable up to a function composition $\h = \pi_z \circ r_{z}$ of a permutation $\pi_z: [K] \rightarrow [K]$ and element-wise invertible transformation $r_{z}: \R^K \rightarrow \R^K$. Or equivalently, the same follows for the generative function $\g \circ \h = \hatg$.
	
	\paragraph{Theorem 2 (Identifiability of the dynamical function $\f$):} Under assumptions \textit{(A0, A1, A2, A3, A4, A5)}, the process noise $\s_t = \bk(\hats_t)$ is identifiable up to a function composition $\bk = \pi_s \circ r_{s}$ of a permutation $\pi_s: [K] \rightarrow [K]$ and an element-wise invertible transformation $r_{s}: \R^K \rightarrow \R^K$. Equivalently, the dynamical function $\h^{-1}_{\texttt{aug}} \circ \faug \circ \kaug = \hatfaug$ is identifiable up to a function composition $\kaug = [\bk, \h] = \pi \circ r$ of a permutation $\pi = [\pi_s, \pi_z]: [2K] \rightarrow [2K]$ and an element-wise invertible transformation $r = [r_{s}, r_{z}] : \R^{2K} \rightarrow \R^{2K}$, where \textbf{Theorem 1} already proves that $\h^{-1}_{\texttt{aug}} = [\h^{-1}, \h^{-1}]$ is an invertible element-wise transformation. 
	
	\section{Practical implementation using variational inference}
	\label{sec:practical}
	We turn our theoretical framework into a practically usable implementation using variational inference, which approximates the true posterior over the noise and the states given the observations. For space considerations, we defer the details of our inference setup to \cref{sec:app:vi}. Please see the below algorithm for implementation details and Figure~\ref{fig:app:arch} for architecture details.
	
	\begin{algorithm}
		{\caption{Practical learning algorithm}}%
		{%
            Requires: Variational posterior networks (\texttt{ICEncoder} and \texttt{NoiseEncoder}) and \texttt{Decoder} 
			\begin{enumerate}
				\item Encode initial condition parameters: $\mu_{\z_0}, \log \sigma^2_{\z_0} = \texttt{ICEncoder}(\x_{1:T_{\text{ic}}})$
                \item Sample initial condition: $\z_0 \sim \mathcal{N}(\mu_{\z_0}, \sigma^2_{\z_0}\mathbf{I})$
				\item For $t \in 1, \ldots, T$:
				\begin{enumerate}
					\item Encode noise parameters: $\mu_{\s_t}, \log \sigma^2_{\s_t} = \texttt{NoiseEncoder}(\x_{1:t}, \z_{t-1})$
                    \item Sample noise: $\s_t \sim \mathcal{N}(\mu_{\s_t}, \sigma^2_{\s_t}\mathbf{I})$
					\item Compute the next latent state: $\z_t = \f(\s_t,\z_{t-1})$
					\item Decode: $\x_t = \texttt{Decoder}(\z_t)$
				\end{enumerate}
				\item Compute \textsc{Elbo}: $\elbo = \elboR - \beta \elboKL$. Samples $\{\s_{1:T}, \z_{0:T}\}$ are used to approximate $\elboKL$: 
				{ \small
                \begin{align*}
                    \elboR &= \sum_{t=1}^T \EX_{\qph(\z_{t}|\dots)}[\log \pth(\x_{t} | \z_{t})], \\
					\elboKL &= D_\text{KL}(\qph(\z_{0} | \x_{1:T}, \bu) \| \pth(\z_{0})) + \sum_{t=1}^T \EX_{q_{\phi}(\z_{t-1} | \dots)}[D_\text{KL}(\qph(\s_{t} | \z_{t-1}, \x_{1:t}, \bu) \| \pth(\s_{t} | \bu))]
				\end{align*}
                }
                \item Update the parameters $\{\theta, \phi\}$.
			\end{enumerate}
		}%
	\end{algorithm}

	\paragraph{Initial encoding.} We map each observation $\x_t$ to an initial embedding $\br_t$ via an \textsc{Mlp} or \textsc{Cnn} depending on the input modality. Using the first $T_\text{ic}=4$ initial embeddings $\br_{1:T_\text{ic}}$, an initial condition encoder (\textsc{Mlp}) outputs the parameters of the variational posterior $q(\z_0 | \x_{1:T_\text{ic}})$ for the initial condition $\z_0$. Our ablations showed that the model is rather insensitive to $T_\text{ic}$.
	
	\paragraph{Sequence prediction.} We start by sampling from the initial value distribution $\z_0 \sim q(\z_0 | \x_{1:T_\text{ic}})$. 
    Next, a forward sequential layer (\textsc{Rnn} + \textsc{Mlp}) takes the initial embeddings $\br_{1:t}$ up to time $t$ and the (sampled) previous latent state $\z_{t-1}$ as input $[\br_{1:t}, \z_{t-1}]$, and outputs the parameters of the variational posterior $q(\s_t | \x_{1:t}, \z_{t-1})$ for the noise variables $\s_{1:T}$. For example, for the first noise variable $\s_1$, the variational posterior is of the form $q(\s_1 | \x_{1}, \z_{0})$.
    Subsequently, given a sample from the noise variable $\s_1 \sim q(\s_1 | \z_0, \x_{1})$ and the sampled initial state $\z_0$, we predict the next state $\z_1 \equiv \f(\z_0,\s_1)$, or more specifically: $z_{k1} \equiv f_k(\z_0, s_{k1})$ for $k \in 1, \ldots, K$. We model each output $k$ of the transition function $f_k$ as a separate \textsc{Mlp}, to encourage the conditional independence of the latent states. We recursively compute the trajectory $\z_{2:T}$ of future latent states and noise posteriors $q(\s_{2:T})$.
	
	\paragraph{Priors and ELBO computation.} We assume a standard Gaussian prior for the initial condition $p(\z_0)=\N(0,I)$. The prior $p(\s_t|\bu) = \prod_k p(s_{kt} | \bu)$ over the noise variables are $1D$ trainable conditional flows. To allow for multi-modal prior distributions for $1D$ noise variables $s_{kt}$, we choose neural spline flows as the flow architecture \citep{durkan2019neural,stimper2023normflows}. For the computation of the KL divergence, the terms containing a conditional flow do not have closed form solutions. We compute them by a Monte Carlo approximation using the sequence samples $\{\z_{0:T}, \s_{1:T}\}$. Finally, the decoder $\mathbf{d}$ outputs the mean of our Gaussian observation model with fixed variance: $p(\x_t|\z_t)=\N(\x_t;\mathbf{d}(\z_t),I)$.
 
\section{Experiments}
\label{sec:experiments}

In this section, we perform experiments on two temporal datasets where the temporal dependencies are governed by an underlying latent dynamical system: (i) a synthetic dataset containing multivariate time-series data similar to \citep{hyvarinen2016unsupervised,hyvarinen2017nonlinear,hyvarinen2019nonlinear,yao2021learning, yao2022temporally}, and (ii) a more challenging reinforcement learning environment composed of high-dimensional video sequences of a physical cartpole system \citep{huang2021adarl,yao2022temporally}. Our goal is to answer the following research questions:
\begin{itemize}
    \item [\textit{(Q1)}] Can our model identify the underlying latent dynamics, which contains three components: latent states $\z_{1:T}$, the transition function $\f$, the process noise $\s_{1:T}$?,
    \item [\textit{(Q2)}] What is the advantage of our model which jointly learns the latent variables and the transition function, compared to a plug-in method which first identifies the latent variables and simply fits a transition function on those?
    \item [\textit{(Q3)}] Ablation: How does each model component contribute to the disentanglement performance?
    \item [\textit{(Q4)}] By learning an identifiable transition function $\f$, can we predict the system state for longer horizons than the training sequence length?
    \item [\textit{(Q5)}] Is identifiable dynamics useful for sample efficient adaptation of dynamical systems?
\end{itemize}

\paragraph{Metrics.} We evaluate the performance on the latent dynamics identification using mean correlation coefficient (\mcc), the standard metric in the nonlinear \ica{} literature \citep{hyvarinen2019nonlinear,yao2021learning}. 
Let us denote the latent states and the process noise for first \Ttrain{} steps by $\ztrain = \z_{0:\Ttrain}$ and $\strain = \s_{1:\Ttrain}$ respectively.
For \textit{(Q1,Q3)}, we measure the validation \mcc{} for the latent sequence \ztrain: $\mccz$. For the synthetic data set, we additionally measure the validation \mcc{} for the noise sequence \strain: $\mccs$. 
For \textit{(Q2,Q4,Q5)}, we measure the future prediction performance.
For this, we take the subsequent \Tfuture{} steps coming after the first \Ttrain{} steps.
We denote the future observations and latent states for these frames as $\xfuture = \x_{\Ttrain+1:}$ and $\zfuture = \z_{\Ttrain+1:}$ respectively.
As metrics, we measure the mean squared error (\mse{}) on the future observations \msexfuture{} and the \mcc{} on the future latent states \mcczfuture{}. 
When \Tfuture{} takes different values, e.g., $\Tfuture{} \in \{2,4,8\}$, we denote the future metrics by \msexfutureh{2}, \msexfutureh{4} and \msexfutureh{8}.
 
\paragraph{Baseline methods.} For \textit{(Q1)}, we compare with nonlinear \ica methods assuming: (i) independent latent processes $\z$ with no temporal or nonstationary structure: \betavae \citep{higgins2018towards}, (ii) mutually independent latent processes $z_{kt} = f_k(z_{k,t-1}, s_{kt})$ with nonstationarity: \tcl \citep{hyvarinen2016unsupervised} and i\textsc{Vae} \citep{khemakhem2020variational}, (iii) mutually independent latent processes $z_{kt} = f_k(z_{k,t-1}, s_{kt})$ with temporal structure: \pcl \citep{hyvarinen2017nonlinear} and \slowvae \citep{klindt2020towards}, and (iv) temporally mixed latent processes $z_{kt} = \f(\z_{t-1}, s_{kt})$ with nonstationarity: two versions of \leap \citep{yao2021learning} having linear and nonparametric transition functions \leaplin and \leapnp. For \textit{(Q2)}, we construct a baseline  based on the closest methods to ours: \leapnp. We first identify the latent variables using \leapnp, and then fit an \textsc{Mlp} on them as the transition function in an offline manner. For \textit{(Q4,Q5)}, we compare with a disentangled deep state-space model \kvae \citep{fraccaro2017disentangled}, a nonlinear \ica method \leaplin with linear transitions that can predict the future states, and the constructed baseline \leapnpmlp.

\subsection{Synthetic experiment}

\begin{table}
\label{table:synth}
\caption{Synthetic experiment results (mean std.dev.~across 5 runs). For methods that cannot predict the future, we leave the $\msexfutureh{\cdot}$ rows empty (N/A). Likewise, we compute $\mccs$ only for \leap and our method as others do not maintain process noise variables.}
  \centering
  \scriptsize
  \begin{tabular}{lccccccccc}
    \toprule
     & \multicolumn{9}{c}{\textsc{Models}} \\
    \cmidrule(r){2-10}
    \textsc{Metrics} & \betavae & \pcl & \tcl & \ivae & \slowvae & \kvae & \leaplin & \leapnp & \textsc{Ours} \\
    \midrule
    \multirow{2}{*}{$\mccz$ \hfill $\uparrow$} & $0.60$ & $0.57$ & $0.39$ & $0.58$ & $0.41$ & $0.64$ & $0.68$ & $0.89$
    &$0.95$ \\
     & \std{0.05} & \std{0.05} & \std{0.07} & \std{0.06} & \std{0.05} & \std{0.05} & \std{0.03} & \std{0.04}
    & \std{0.08} \\
    \midrule
    \multirow{2}{*}{$\mccs$ \hfill $\uparrow$} & \multirow{2}{*}{N/A} & \multirow{2}{*}{N/A} & \multirow{2}{*}{N/A} & \multirow{2}{*}{N/A} & \multirow{2}{*}{N/A} & \multirow{2}{*}{N/A} & $0.14$ & $0.26$ & $0.66$ \\
    &  &  &  &  &  &  & \std{0.01} & \std{0.04} & \std{0.09} \\
    \midrule
    \multirow{2}{*}{\msexfutureh{2} \hfill$\downarrow$} & \multirow{2}{*}{N/A} & \multirow{2}{*}{N/A} & \multirow{2}{*}{N/A} & \multirow{2}{*}{N/A} & \multirow{2}{*}{N/A} & $1.27$ & $0.22$ & \multirow{2}{*}{N/A} & $0.06$ \\
    &  &  &  &  & & \std{0.19} & \std{0.03} & & \std{0.01} \\
    \midrule
    \multirow{2}{*}{\msexfutureh{4} \hfill $\downarrow$} & \multirow{2}{*}{N/A} & \multirow{2}{*}{N/A} & \multirow{2}{*}{N/A} & \multirow{2}{*}{N/A} & \multirow{2}{*}{N/A} & $1.32$ & $0.18$ & \multirow{2}{*}{N/A} & $0.09$ \\
    &  &  &  &  & & \std{0.27} & \std{0.03} & & \std{0.01} \\
    \midrule
    \multirow{2}{*}{\msexfutureh{8} \hfill $\downarrow$} & \multirow{2}{*}{N/A} & \multirow{2}{*}{N/A} & \multirow{2}{*}{N/A} & \multirow{2}{*}{N/A} & \multirow{2}{*}{N/A} & $1.72$ & $0.59$ & \multirow{2}{*}{N/A} & $0.21$ \\
    &  &  &  &  & & \std{0.82} & \std{0.13} & & \std{0.03} \\
    \bottomrule
  \end{tabular}
\end{table}

\begin{figure}
    \centering
    \includegraphics[width=\linewidth,trim={0cm 71.5cm 30cm 0},clip]{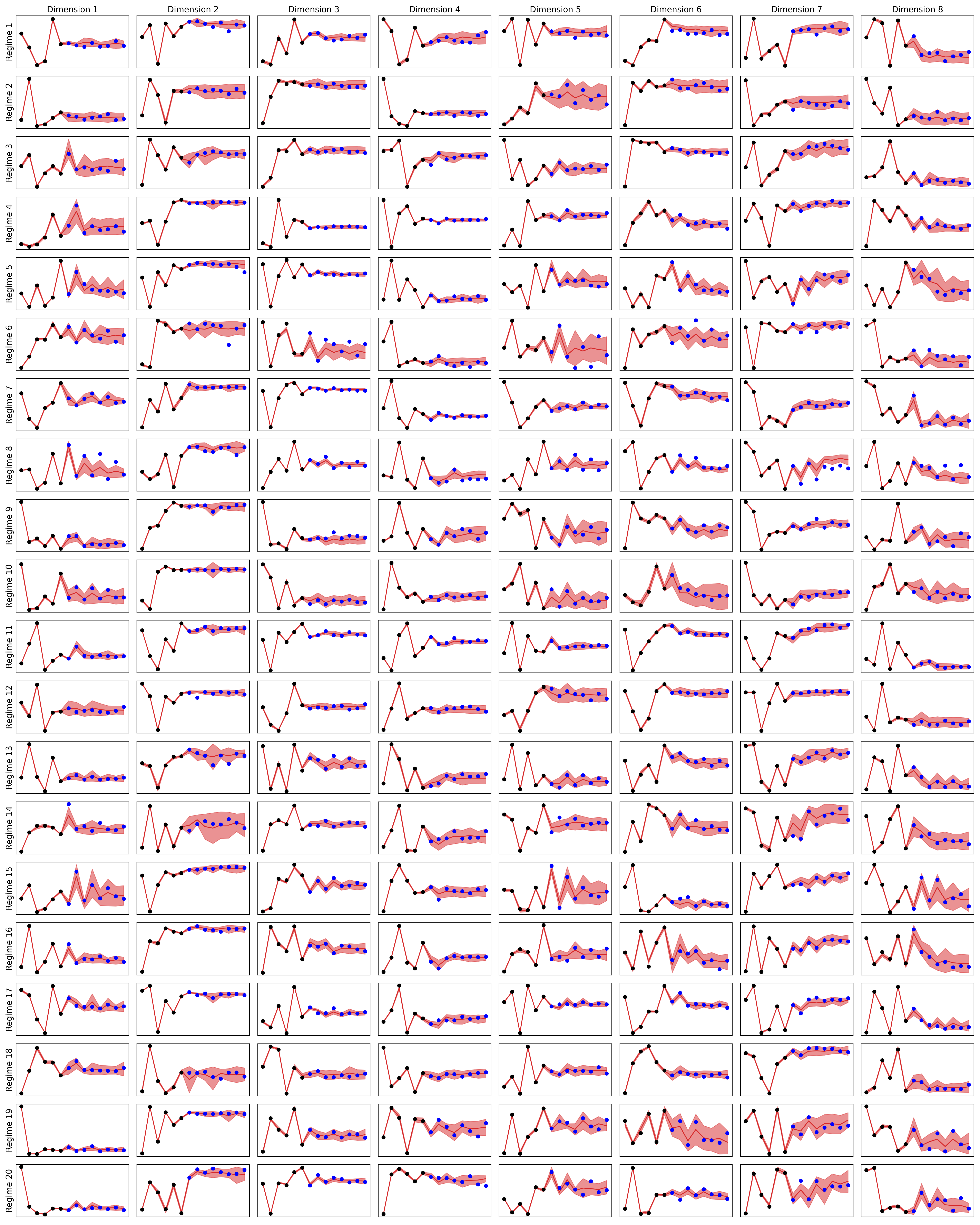}
    \includegraphics[width=\linewidth,trim={0cm 19cm 30cm 53.5cm},clip]{figs/uncertainty_plot.png}
    \caption{Model predictions in the data space with the estimated uncertainties. Our model uses the first $T_0=2$ data points to encode the initial latent state $\z_0$ and $\Tdyn=4$ data points to encode the noise sequence \strain. We unroll our model for $T_0+\Tdyn+\Tfuture=14$ steps ahead, where the future noise variables $\s_{7:14}$ follow the learned prior flow. We draw 32 trajectory samples $\x_{0:14}$ by sampling from the initial state and process noise. Above, black and blue dots show the training and test data points. The red curves are the mean trajectories and the red region corresponds to $\pm 2$ standard deviation computed empirically. We observe near-perfect predictions and low uncertainty for the input data (the first $\Ttrain=T_0+\Tdyn = 6$ time points) while the uncertainty grows as we unroll over time. Further, the uncertainty grows even more when the model predictions are off. Therefore, almost all test points lie in the $\pm 2$ std region, reflecting the high calibration level our probabilistic model attains. Please see Figure~\ref{fig:app:uncertainty} for all regimes.}
    \label{fig:uncertainty}
\end{figure}

As common in nonlinear \ica literature \citep{hyvarinen2019nonlinear, morioka2021independent, yao2021learning, yao2022temporally}, we first set up a synthetic experiment to show that \textit{(Q1)} our model recovers the latent dynamics, and hence \textit{(Q4)} achieves a higher future prediction accuracy. On the same synthetic setup, we \textit{(Q3)} analyze how each model component affects the performance in an ablation study, and show that \textit{(Q2)} our method predicts the future behavior better than a plug-in method which trains a transition function on top of the identified latent states.

\paragraph{Dataset.} Similar to \citet{yao2021learning,yao2022temporally}, we set up a synthetic data experiment containing multivariate time-series. We set the dimension of $\s$, $\z$ and $\x$ to $K=8$. Same as \citet{yao2021learning,yao2022temporally}, we use a 2-linear layer random MLP as the generative function $\g$, 2-linear layer random MLP as the transition function $\f$ and we choose $\z_{1:T}$ to be a second-order Markov process, i.e., $\z_t = \f(\z_{t-2}, \z_{t-1}, \s_t)$. The number of environments is $R=20$. For each environment, we generate 7500/750/750 sequences as train/validation/test data following our generative model. The distribution of the process noise $\s_t$ is conditioned on the environment index $\bu$. Each sequence has length $T=T_0 + \Tdyn + \Tfuture = 2+4+8=14$. As we have a second-order Markov process, first $T_0=2$ states are spared as initial states. The next 4 observations $\x_{1:4}$ are used for training the dynamical model. The last 8 observations $\x_{5:12}$ are used for assessing the performance of future estimation. We choose the future prediction horizon $T_{\text{future}}=8$ as the double of the training sequence length. If the dynamics are truly identified, the model should predict future states well, even for a longer horizon.

\paragraph{Main results.} $\mccz$, $\mccs$ and $\msexfuture$ results are shown in Table 1. Our model recovers latent states $\z$ and process noise $\s$ better than the baselines, as demonstrated by a higher correlation with the true latent states and the true process noise (in terms of $\mccz$ and $\mccs$). This leads to a higher accuracy in future prediction, in terms of $\msexfuture$ with prediction horizons $\{2,4,8\}$. Notice that 8-step future prediction corresponds to the double the amount of dynamics steps the models see during the training. As the prediction horizon increases, we see that the difference in the future prediction performance between our method and the baselines also increases.

\begin{wrapfigure}{r}{4.8cm}
    \centering
    % \vspace{-0.4cm}
    \includegraphics[width=1.0\linewidth]{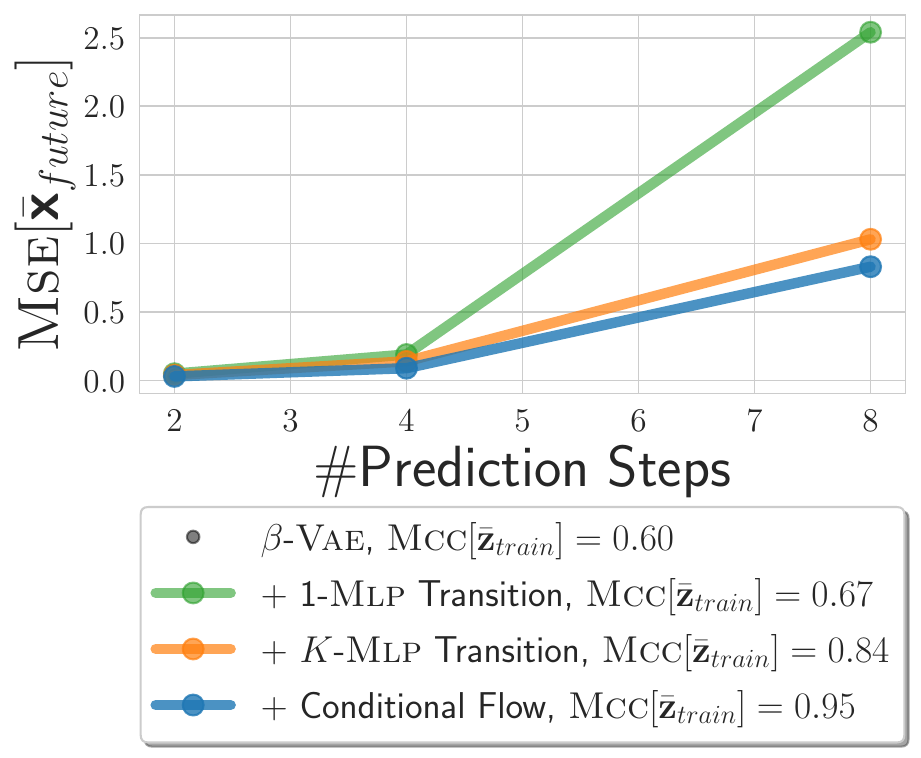}
    \caption{\mcc{} vs.~\mse{} results for ablations for different time steps.}
    \label{fig:leap-two-stages}
    \label{fig:ablation}
    % \vspace{-0.4cm}
\end{wrapfigure}
\paragraph{Ablation study.} To understand how each model component affects the identification and prediction performance \textit{(Q3)}, we set up an ablation study. We chose \betavae{} \citep{higgins2018towards} as the base model. On top of it, we add three model components one by one: (i) {\color{green!50!black} \mlpdynone}: a single \textsc{Mlp} modeling $K$ outputs of the transition function $\{f_k\}_{k=1}^K$, (ii) {\color{orange!90!black}\mlpdynK}: $K$ independent \textsc{Mlp}s modeling $K$ outputs of the transition function $\{f_k\}_{k=1}^K$, and (iii) {\color{blue}\textsc{Ours}}: an additional conditional normalizing flow to model the nonstationarity $p(\s | \bu)$. We present $\msexfuture$ results for prediction steps $T_{\text{pred}} \in \{2, 4, 8\}$ in \cref{fig:ablation}, together with the $\mccz$ values in the legend. We see that using $K$ different \textsc{Mlp}s and a conditional normalizing flow have similar performance improvements for disentanglement. More importantly, the results suggest that as we predict for longer horizons (e.g. 8 steps), a higher \mcc{} on the latent states lead to a larger improvement on the future prediction performance. Finally, in \cref{fig:uncertainty}, we visualize our model predictions in the data space, highlighting the calibrated uncertainty handling of our probabilistic approach.

\begin{wrapfigure}{r}{6.5cm}
    \centering
    % \vspace{-0.6cm}
    \includegraphics[width=1.0\linewidth]{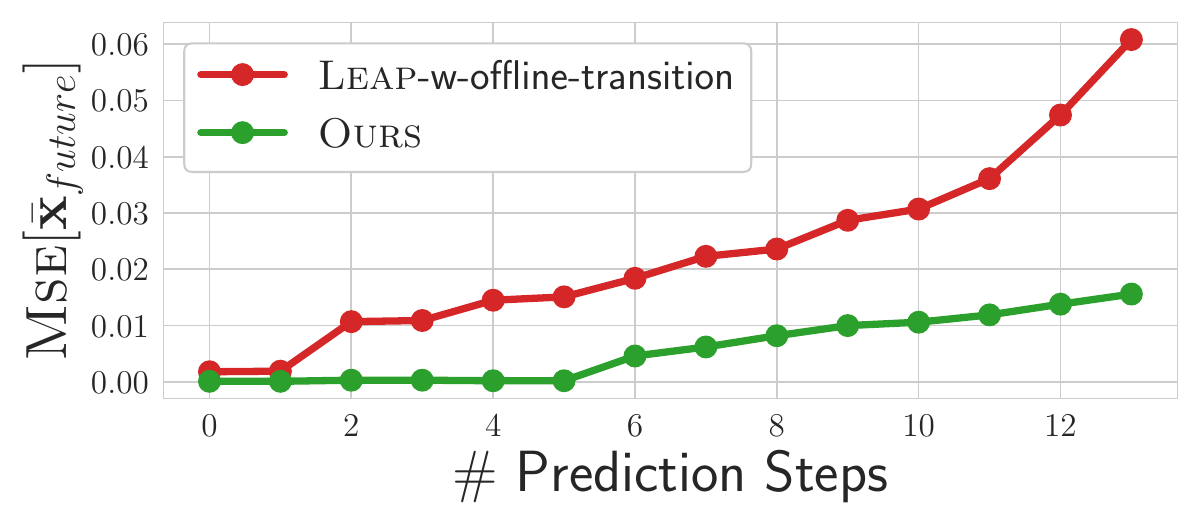}
    \caption{A comparison of MSEs achieved by our approach and two-stage LEAP training.}
    \label{fig:leap-two-stages}
    % \vspace{-0.4cm}
\end{wrapfigure}
\paragraph{\leap with offline transition training vs.~our approach.} Next, we demonstrate the benefits of our learning procedure, which identifies the transition and generative functions jointly.
In principle, one can use an off-the-shelf identifiable representation learning method, such as \leapnp, identify the ground-truth factors, and fit a state transition function on the inferred latent codes. 
Such a transition function could approximate the unknown transition function equally well.
We implement and test this two-stage procedure on our synthetic dataset. 
To learn the unknown transitions, we train a three-layer MLP with leaky relu activations on the latent state sequence \ztrain{} inferred by \leapnp (different hyperparameters yielded similar results).
We unroll the learned transition function for an additional \Tfuture{} time steps, decode all the latent states, and compute the \mse{} in the data space.
We repeat this three times and compare the average error with the error of our approach in Figure~\ref{fig:leap-two-stages}.
Our approach achieves smaller errors at all time points, and the gap widens as we predict for longer horizons.
This indicates that identifying the generative and transition functions jointly results in a better model.

\subsection{Modified Cartpole experiment}

Next, we experiment on challenging video data of a physical cartpole system with nonlinear dynamics \citep{huang2021adarl,yao2022temporally}. Similarly to the synthetic experiment, we show our method yields more accurate latent system identification and future predictions than baselines \textit{(Q1,Q4)}.

\paragraph{Cartpole dataset.} We use the modified cartpole setting in \citep{huang2021adarl,yao2021learning}, an adaptation of the environment in OpenAI Gym \citep{brockman2016openai}. The environment simulates a physical system with a pole attached to a cart from a pivot point. The system has 4 states: position and velocity of the cart, and the angle and the angular velocity of the pole ($[x_{\text{cart}}, \theta_{\text{pole}}, \dot{x}_{\text{cart}}, \dot{\theta}_{\text{pole}} ]$). Observations consist of video sequences of the system and we use the true states only to compute \mcc{}.
The setup has 5 source domains used for training with the different levels of gravity $g = \{5,10,20,30,40\}$ and mass $m=1.0$. 
We use a target domain with $g=90$ to check the methods' ability to \textit{extrapolate}. 
We observe the system under random binary actions, which move the cart to left or right. 
We generate sequences of length $T=T_0 + \Tdyn + \Tfuture = 1+8+8=17$, the first frame for the initial state, the next 8 frames for training the model, and the last 8 frames for future prediction. For each source domain, we have 900/100 sequences as train/validation data. For the target domain, we have data sets with different numbers of samples $N_{\text{target}} = \{20, 50, 1000\}$.

\paragraph{Main results.} 
\cref{tab:cartpole} presents accuracies of reconstruction and prediction of the latent states. (Here also predictions are compared in the latent space since the true future states contain pole angles unseen during training, resulting in blurry reconstructions.)
As expected, methods that take complex temporal dependencies into account (\leaplin, \leapnp and \textsc{Ours}) perform better than other baselines. Our model recovers latent states the best, i.e., the inferred states are more strongly correlated with the ground-truth.
Also, in line with the synthetic experiment, more accurate latent states on the initial training steps also lead to higher future prediction accuracy.

\paragraph{Adaptation experiment.} 
Next, we take the converged checkpoints of \leaplin and our approach and adapt them to the target domain described earlier. 
Because both models claim to identify the latent states consisting of positions and velocities, we expect them to transfer well to a new dataset with the same generative mechanism.
We consider a domain shift ($g=40 \rightarrow 90$) and observe that the \mcc values for both models drop to the range $[0.60-0.65]$.
To mitigate this, we train the models on a small ``adaptation dataset'' with $N=\{20,50,1000\}$ sequences. 
We try adapting (i) only the dynamics (including process noise flow) and (ii) the entire network.
\cref{fig:app:adaptation} shows the findings.
First, despite decreasing the loss, \leaplin \mcc score never improves irrespective of the dataset size.
Importantly, our model improves its \mcc by about $25\%$ with only $50$ adaptation trajectories.
Interestingly, optimizing only the dynamics results in better MCC than optimizing all modules. 

\begin{table}
\caption{Results for the cartpole experiment. N/A: the method cannot predict future.
}
\label{tab:cartpole}
  \centering
  \small
  \begin{tabular}{lcccccc}
    \toprule
     & \multicolumn{6}{c}{\textsc{Models}} \\
    \cmidrule(r){2-7}
    \textsc{Metrics} & \pcl & \slowvae & \kvae & \leaplin & \leapnp & \textsc{Ours} \\
    \midrule
    $\mccz$ & $0.62$ &  $0.54$ & $0.60$ & $0.83$ & $0.80$ & $0.95$ \\
    $\mcczfutureh{8}$ & N/A & N/A & $0.55$ & $0.71$ & N/A & $0.93$ \\
    \bottomrule
  \end{tabular}
\end{table}

\section{Discussion}
We have presented the first latent dynamical system that allows for the identification of the unknown transition function, and theoretically proved its identifiability, based on standard assumptions. We evaluated our approach on synthetic data and the Cartpole environment and showed that (i) the estimated latent states correlated strongly with the ground truth, (ii) our method had the highest future prediction accuracy with calibrated uncertainties, and (iii) it could adapt to new environments using a handful of data. The main limitation stems from the identifiability assumptions adopted. For future work, it would be intriguing to examine their violation might influence the final model's performance. Further, our most realistic empirical demonstration is based on data from a physics simulator, i.e., we leave studies on real-world datasets for future. Finally, demonstrating improved downstream performance from our method, e.g. in model-based policy learning, would be of interest.

	\bibliography{ms}
	
	\newpage

    \appendix
 
    \section{Identifiability Theory}
    \label{sec:app:identifability}

In this section, we discuss the identifiability of the latent states and the transition function, and provide the detailed proofs.

We assume a latent dynamical system which is viewed as high-dimensional sensory observations $\x_{1:T}$, where $t$ is the time index and $\x_t \in \R^D$. We assume a sequence of latent states $\z_{1:T}$, with $\z_t \in \R^K$, are instantaneously mapped to observations via a generative function $\g: \R^K \rightarrow \R^D$:
\begin{equation}
    \label{app:eq:gen}
    \x_t = \g(\z_t).
\end{equation}
The latent states $\z_{1:T}$ evolve according to Markovian dynamics:
\begin{equation}
    \z_t = \f(\z_{t-1},\s_{t}),
\end{equation}
where $\f: \R^{2K} \rightarrow \R^K$ is an auto-regressive transition function and $\s_t \in \R^K$ corresponds to process noise.

Our aim is to jointly identify the latent states $\z_{1:T}$, the dynamics function $\f$, and the process noise $\s_{1:T}$. We remind that previous works \citep{klindt2020towards, yao2021learning, yao2022temporally, song2023temporally} have concentrated on identifying the latent states $\z_{1:T}$, possibly with linear transition approximations, but not a general transition function $\f$. 
Yet, without a general $\f$, the methods can estimate the underlying states only when corresponding observations are provided or provide simplistic approximations in their absence. Hence, they cannot predict complex future behavior reliably.

Notice that learning a provably identifiable transition function $\f : \R^{2K} \rightarrow \R^K$ is not straightforward, since the transition function is not injective. A naive solution can be to simply use a plug-in method \citep{yao2021learning, yao2022temporally} for identifying the latents and then fitting a transition function $\f$ on the estimated latents, however, we show empirically in our experiments that it leads to poor prediction accuracy for the future behavior.

\subsection{Nonlinear \ica}

The nonlinear \ica assumes that the data is generated from independent latent variables $\z$ with a nonlinear generative function $\g$, following \cref{app:eq:gen}. It is well-known to be non-identifiable for i.i.d.~data \citep{hyvarinen1999nonlinear, locatello2019challenging}. Recent seminal works \citep{hyvarinen2016unsupervised, hyvarinen2017nonlinear, hyvarinen2019nonlinear} showed that \textit{autocorrelation} and \textit{nonstationarity} existent in non-i.i.d.~data can be exploited to identify latent variables in an unsupervised way.
Compared to the vanilla \ica that considers independence only along latent dimensions, the idea of these works is to introduce additional independence constraints reflecting the existent structure in the data. These additional constraints are formulated mathematically as \textit{identifiability assumptions}, which restrict the space of the generative function $\g$ and the space of the latent prior $p_{\z}$ \citep{hyvarinen2023nonlinear,xi2023indeterminacy}. The key insight is that, after sufficiently constraining the latent prior $p_{\z}$ using such assumptions, identifying the latent variables $\z_t$ and identifying the injective generative function $\g$ become equivalent tasks \citep{xi2023indeterminacy}.

\subsection{Augmented dynamics for identifiable systems}
\label{ssec:augmented-dynamics} 

To identify the transition function $\f$ such that $\z_t = \f(\z_{t-1}, \s_t)$, we will use the same insight: After sufficiently constraining the noise prior $p_{\s}$; given an identifiable latent pair $(\z_{t-1}, \z_t)$, identifying the noise variables $\s_t$ and identifying the bijective dynamics function $\f$ should be equivalent. Hence, in addition to the identifiability assumptions restricting the space of the generative function $\g$ and the space of the latent prior $p_{\z}$, we will further restrict the space of the dynamics function $\f$, and the space of the noise prior $p_{\s}$.

First, let us note that the identifiability of the process noise $\s_t$ is not trivial since the dynamics function $\f: \R^{2K} \rightarrow \R^K$ is not an injective function and hence it does not have an inverse. Following the independent innovation analysis (\textsc{Iia}) framework \citet{morioka2021independent}, we trivially augment the image space of the transition function and denote the bijective augmented function by $\faug: \R^{2K} \rightarrow \R^{2K}$:
\begin{align}
    \zpair
    = \faug \left( \fauginpair \right) = \faugimpl
\end{align}
Contrary to \citet{morioka2021independent}, which use an augmented autoregressive model on \textit{observations} $(\x_{t-1}, \x_t)$, our formulation captures the functional dependence between a \textit{latent pair} $(\z_{t-1}, \z_t)$ and the process noise $\s_t$.

Next, we make the standard assumption in the temporal identifiability literature \citep{klindt2020towards, yao2021learning, yao2022temporally, song2023temporally} that each dimension of the transition function $\{f_k\}_{k=1}^K$ is influenced by a single process noise variable $s_{kt}$. The output is a single latent variable $z_{kt}$:
\begin{align}
    z_{kt} &= f_k (\z_{t-1}, s_{kt}), \quad \text{for } k \in 1, \ldots, K \text{ and } t \in 1, \ldots, T.
\end{align}
Notice that this does not impose a limitation on the generative model, it just creates a segmentation between noise variables and latent variables. For example, if this assumption is violated and there exists a noise variable $s_{kt}$ that affects both $z_{it}$ and $z_{jt}$ with $i \neq j$, then the noise variable $s_{kt}$ can instead be modeled as a latent variable $z_{kt}$.

We re-state the full generative model for completeness:
\begin{align}
    \label{eq:gen-ic}
    \z_0 &\sim p_{\z_0}(\z_0), &&\texttt{\# initial state} \\
    \label{eq:gen-noise}
    \s_t &\sim p_{\s | \bu}(\s_t | \bu) = \prod_{k} p_{s_{k} | \bu}(s_{kt} | \bu), \quad \forall t \in 1, \ldots, T, &&\texttt{\# process noise} \\
    \label{eq:gen-transition}
    \zpair &= \faug \left( \fauginpair \right) = \faugimpl, \quad \forall t \in 1, \ldots, T, &&\texttt{\# state transition} \\
    \label{eq:gen-obs}
    \xpair &= \gaug \left( \zpair \right) = \augvec{\g(\z_t)}{\g(\z_{t-1})}, \qquad \forall t \in 2, \ldots, T. &&\texttt{\# observation mapping} 
\end{align}
where $\bu$ is an auxiliary variable, which modulates the noise distribution $p_{\s|\bu}$.

\subsection{Proof of Theorem 1: Identifiability of the latent states $\z_t$}
\label{app:sec:proof1}

This result is already shown in \citep[Appendix~A.3.2]{yao2021learning}. Here, we follow \citet{klindt2020towards,yao2021learning,yao2022temporally} and repeat their results in our notation as we also make use of this result in \cref{app:sec:proof2}.

The injective functions $\g, \hatg: \R^K \rightarrow \R^D$ are bijective between the latent space $\R^K$ and the observation space $\mathcal{X} \subset \R^D$. We denote the inverse functions from the restricted observation space to the latent space by $\g^{-1}, \hatg^{-1}$. This is also implicitly assumed in \citep{klindt2020towards, yao2021learning, yao2022temporally, song2023temporally}.

We have
\begin{align}
    \x_t = \hatg(\hatz_t) = \bigg( (\g \circ \underbrace{\g^{-1}) \circ \hatg}_{\h} \bigg)(\hatz_t) \implies \hatg = \g \circ \h \implies \z_t = \h(\hatz_t).
\end{align}
The function $\h: \hatz_t \mapsto \z_t$ maps the learned latents to the ground-truth latents. To show it is bijective, we need to show it is both injective and surjective. Following \citet{klindt2020towards}, it is injective since it is a composition of injective functions. Assume it is not surjective, then there exists a neighborhood $\mathbf{U_{z}}$ for which $\g(\mathbf{U_{z}}) \notin \hatg(\R^{K})$. This implies that the neighborhood of images generated by $\g(\mathbf{U_{z}})$ has zero density under the learned observation density $p_{\hatgaug, \hatfaug, \hat{p}_{\s | \bu}}(\g(\mathbf{U_{z}})) = 0$, while having non-zero density under the ground-truth observation density $p_{\g, \faug}(\x)$: $p_{\gaug, \faug, p_{\s | \bu}}(\g(\mathbf{U_{z}})) > 0$. This contradicts the assumption that the observation densities match everywhere. Then, $\h$ is surjective.

We perform change of variables on the conditional latent density:
\begin{align}
    \log p(\hatz_t | \hatz_{t-1}, \bu) &= \log p(\z_t | \z_{t-1}, \bu) + \log |\bH_t|, \\
    \sum_{k=1}^K \underbrace{\log p(\hat{z}_{kt} | \hatz_{t-1}, \bu)}_{\hatlogz{k}} &= \sum_{k=1}^K \underbrace{\log p(z_{kt} | \z_{t-1}, \bu)}_{\logz{k}} + \log |\bH_t| \\
    \sum_{k=1}^K \hatlogz{k} &= \sum_{k=1}^K \logz{k} + \log |\bH_t|
\end{align}
where $\bH_t = \mathbf{J_h}(\hatz_t)$ is the Jacobian matrix of $\h$ evaluated at $\hatz_t$. First, we take derivatives of both sides with respect to $\hat{z}_{it}$:
\begin{align}
    \hatlogz{i} &= \sum_{k=1}^K \frac{\partial \logz{k}}{\partial z_{kt}} \frac{\partial z_{kt}}{\partial \hat{z}_{it}} + \frac{\partial \log |\bH_t|}{\partial \hat{z}_{it}}
\end{align}
Second, take derivatives with respect to $\hat{z}_{jt}$:
\begin{align}
    0 &= \sum_{k=1}^K \left( \frac{\partial^2 \logz{k}}{\partial z^2_{kt}} \frac{\partial z_{kt}}{\partial \hat{z}_{it}} \frac{\partial z_{kt}}{\partial \hat{z}_{jt}} + \frac{\partial \logz{k}}{\partial z_{kt}} \frac{\partial z^2_{kt}}{\partial \hat{z}_{it} \partial \hat{z}_{jt}} \right) + \frac{\partial \log |\bH_t|}{\partial \hat{z}_{it} \partial \hat{z}_{jt}}
\end{align}
Lastly, take derivatives with respect to $u_{l}$:
\begin{align}
    0 &= \sum_{k=1}^K \left( \frac{\partial^3 \logz{k}}{\partial z^2_{kt} \partial u_l} \frac{\partial z_{kt}}{\partial \hat{z}_{it}} \frac{\partial z_{kt}}{\partial \hat{z}_{jt}} + \frac{\partial^2 \logz{k}}{\partial z_{kt} \partial u_l} \frac{\partial z^2_{kt}}{\partial \hat{z}_{it} \partial \hat{z}_{jt}} \right), \\
    &= \sum_{k=1}^K \left( \frac{\partial^3 \logz{k}}{\partial z^2_{kt} \partial u_l} [\bH_t]_{ki} [\bH_t]_{kj} + \frac{\partial^2 \logz{k}}{\partial z_{kt} \partial u_l} \frac{\partial z^2_{kt}}{\partial \hat{z}_{it} \partial \hat{z}_{jt}} \right),
\end{align}
since the Jacobian $\bH_t$ does not depend on $\bu$. Using the sufficient variability assumption \textit{(A4)} for the latent states $\z_t$, we can plug in $2K$ values of $\bu_1, \ldots, \bu_{2K}$ for which the partial derivatives of the log conditional density $\logz{k}$ form linearly independent vectors $\bv(\z_t, \bu)$. We see that the coefficients of these linearly independent vectors have to be zero: $[\bH_t]_{ki} [\bH_t]_{kj} = 0$. This implies that the Jacobian matrix $\bH_t$ of the transformation $\z_t = \h(\hatz_t)$ has at most $1$ nonzero element in its rows. Therefore, the learned latents $\hatz_t$ are equivalent to the ground-truth latents $\z_t$ up to permutations and invertible, element-wise nonlinear transformations.

\subsection{Proof of Theorem 2: Identifiability of the latent transition $\f$}
\label{app:sec:proof2}

Here, we prove our main theoretical contribution, \textbf{Theorem 2}. We start by writing the generative process for the observation pair $(\x_{t-1}, \x_t)$:
\begin{align}
    \xpair &= (\hatgaug \circ \hatfaug) \left( \augvec{\hats_t}{\hatz_{t-1}} \right) = \augvec{(\hatg \circ \hatf)(\hatz_{t-1}, \hats_t)}{\hatg(\hatz_{t-1})} \\
    \label{app:eq:proof2-aug}
    &= (\gaug \circ \faug) \circ \underbrace{(\gaug \circ \faug)^{-1} \circ (\hatgaug \circ \hatfaug)}_{\kaug} \left( \augvec{\hats_t}{\hatz_{t-1}} \right)
\end{align}
The function $\kaug: \R^{2K} \rightarrow \R^{2K}$ maps the learned pair $(\hats_t, \hatz_{t-1})$ to the ground-truth pair $(\s_t, \z_{t-1})$:
\begin{align}
    \augvec{\s_t}{\z_{t-1}}
    &= \kaug \left( \augvec{\hats_t}{\hatz_{t-1}} \right) = \augvec{\bk_1(\hats_t, \hatz_{t-1})}{\bk_2(\hats_t, \hatz_{t-1})}.
\end{align}
Similar to the function $\h$ being bijective, it follows that $\kaug$ is also bijective.

Now, we want to show that the augmented function $\kaug$ decomposes into invertible block-wise functions $\bk_1, \bk_2: \R^K \rightarrow \R^K$ such that (i) $\bk_1$ only depends on $\hats_t$, i.e., $\s_t = \bk_1(\hats_t)$ and (ii) $\bk_2$ only depends on $\hatz_{t-1}$, i.e., $\z_{t-1} = \bk_2(\hatz_{t-1})$. It is easy to show (ii), since $\bk_2 = \id_{\z} \circ \g^{-1} \circ \hatg \circ \id_{\hatz} = \h$ and we have already shown in \cref{app:sec:proof1} that the function $\h: \hatz_t \mapsto \z_t$ is equal to $\h = \g^{-1} \circ \hatg$ and bijective.
\begin{align}
    \augvec{\s_t}{\z_{t-1}}
    &= \kaug \left( \augvec{\hats_t}{\hatz_{t-1}} \right) = \augvec{\bk_1(\hats_t, \hatz_{t-1})}{\h(\hatz_{t-1})}.
\end{align}
To show (i), we start by performing change of variables for the transformation $\kaug: (\hats_t, \hatz_{t-1}) \mapsto (\s_t, \z_{t-1})$:
\begin{align}
    \log p(\hats_t, \hatz_{t-1} | \bu) &= \log p(\kaug(\hats_t, \hatz_{t-1}) | \bu) + \log |\mathbf{J_{\kaug}}(\hats_t, \hatz_{t-1})|, \\\
    \label{app:eq:proof2-l1}
    &= \log p([\bk_1(\hats_t, \hatz_{t-1}), \h(\hatz_{t-1})] | \bu) + \log |\mathbf{J_{\kaug}}(\hats_t, \hatz_{t-1})|,
\end{align}
where $\mathbf{J_{\kaug}}(\hats_t, \hatz_{t-1})$ is the Jacobian matrix for the augmented function $\kaug$ evaluated at $(\hats_t, \hatz_{t-1})$. As the process noise is temporally independent given $\bu$, $\hats_t \indep \hatz_{t-1} | \bu$ and $\s_t \indep \z_{t-1} | \bu$, we factorize the densities in \cref{app:eq:proof2-l1}:
\begin{align}
    \log p(\hats_t | \bu) + \log p(\hatz_{t-1} | \bu) &= \log p(\bk_1(\hats_t, \hatz_{t-1}) | \bu) + \log p(\h(\hatz_{t-1}) | \bu) + \log |\mathbf{J_{\kaug}}(\hats_t, \hatz_{t-1})|.
\end{align}
The Jacobian $\mathbf{J_{\kaug}}$ is upper block-diagonal since $\z_{t-1}$ does not depend on $\hats_t$: $\mathbf{J_{\kaug}} = \begin{bmatrix}
    \frac{\partial \s_t}{\partial \hats_t} & \mathbf{*} \\ \mathbf{0} & \bH_t
\end{bmatrix}$ and its log determinant factorizes $\log |\mathbf{J_{\kaug}}(\hats_t, \hatz_{t-1})| = \log|\bH_t| + \log |\frac{\partial \s_t}{\partial \hats_t}|$:
\begin{align}
    \log p(\hats_t | \bu) + \log p(\hatz_{t-1} | \bu) &= \log p(\bk_1(\hats_t, \hatz_{t-1}) | \bu) + \log p(\h(\hatz_{t-1}) | \bu) + \log|\bH_t| + \log |\frac{\partial \s_t}{\partial \hats_t}|.
\end{align}
In addition, the noise is conditionally independent over its dimensions given $\bu$. Therefore, we can further factorize the densities $p(\hats_t | \bu) = \prod_k p(\hat{s}_{kt} | \bu)$ and $p(\bk_1(\hats_t, \hatz_{t-1}) | \bu) = p(\s_t | \bu) = \prod_k p(s_{kt} | \bu)$ with $\s_t = \bk_1(\hats_t, \hatz_{t-1})$:
\begin{align}
    \sum_k \underbrace{\log p(\hat{s}_{kt} | \bu)}_{\hat{q}_k(\hat{s}_{kt}, \bu)} + \log p(\hatz_{t-1} | \bu) &= \sum_k \underbrace{\log p(s_{kt} | \bu)}_{q_{k}(s_{kt}, \bu)} + \log p(\h(\hatz_{t-1}) | \bu) + \log|\bH_t| + \log |\frac{\partial \s_t}{\partial \hats_t}|,
\end{align}
We take the derivative of both sides with respect to $\hat{s}_{it}$:
\begin{align}
    \frac{\partial \hatlogs{i}}{\partial \hat{s}_{it}} &= \sum_k \frac{\partial q_{k}(s_{kt}, \bu)}{\partial s_{kt}} \frac{\partial s_{kt}}{\partial \hat{s}_{it}}+ \frac{\partial \log |\frac{\partial \s_t}{\partial \hats_t}|}{\partial \hat{s}_{it}}.
\end{align}
Next, we take the derivative with respect to $u_l$ with $l$ being an arbitrary dimension:
\begin{align}
    \label{eq:idf-res-minus1}
    \frac{\partial^2 \hatlogs{i}}{\partial \hat{s}_{it} \partial u_l} &= \sum_k \frac{\partial^2 q_{k}(s_{kt}, \bu)}{\partial s_{kt} \partial u_l} \frac{\partial s_{kt}}{\partial \hat{s}_{it}},
\end{align}
since $|\frac{\partial \s_t}{\partial \hats_t}|$ does not depend on $\bu$. Lastly, take the derivative of both sides with respect to $\hat{z}_{j,t-1}$:
\begin{align}
    \label{eq:idf-res}
    0 &= \sum_k \left( \frac{\partial^3 q_{k}(s_{kt}, \bu)}{\partial s^2_{kt} \partial u_l} \frac{\partial s_{kt}}{\partial \hat{s}_{it}} \frac{\partial s_{kt}}{\partial \hat{z}_{j,t-1}} + \frac{\partial^2 q_{k}(s_{kt}, \bu)}{\partial s_{kt} \partial u_l} \frac{\partial^2 s_{kt}}{\partial \hat{s}_{it} \partial \hat{z}_{j,t-1}} \right).
\end{align}
Inspecting the \cref{eq:idf-res}, to ensure the sufficient variability assumption \textit{(A5)} for the process noise $\s_t$, the term $\frac{\partial s_{kt}}{\partial \hat{s}_{it}} \frac{\partial s_{kt}}{\partial \hat{z}_{j,t-1}} = 0$. Following a similar reasoning with \citet{morioka2021independent}, this implies that any dimension of $\s_t$ does not depend on $\hats_t$ and $\hatz_{t-1}$ at the same time. Since $\s_t \indep \z_{t-1} | \bu$ and $\z_{t-1} = \h(\hatz_{t-1})$, $\s_t$ has to depend solely on $\hats_{t}$: $\s_t = \bk_1(\hatz_{t-1}, \hats_t) = \bk(\hats_t)$. We conclude that the augmented function $\kaug$ decomposes into invertible block-wise functions $\bk$ and $\h$: $\kaug = [\bk, \h]$

Now, let's get back to \cref{eq:idf-res-minus1}. Denote the Jacobian matrix of function $\bk$ by $\mathbf{J_k}$ and its evaluation at $\hats_t$ by $\mathbf{J_k}(\hats_t) = \K_t$. Take the derivative of both sides with respect to $\hat{s}_{mt}$ for some index $m$:
\begin{align}
    \label{app:eq:proof2-l2}
    0 &= \sum_k \left( \frac{\partial^3 q_{k}(s_{kt}, \bu)}{\partial s^2_{kt} \partial u_l} [\K_t]_{ki} [\K_t]_{km} + \frac{\partial^2 q_{k}(s_{kt}, \bu)}{\partial s_{kt} \partial u_l} \frac{\partial^2 s_{kt}}{\partial \hat{s}_{it} \partial \hat{s}_{mt}} \right).
\end{align}
Inspecting the \cref{app:eq:proof2-l2}, we see that to ensure the sufficient variability assumption \textit{(A5)}, the product $[\K_t]_{ki} [\K_t]_{km} = 0$. This implies that each dimension $s_{kt}$ of the true latent state depends only on a single dimension of the learned process noise $\hats_t$. Hence, the function $\bk$ is equal to a composition of permutation and element-wise, invertible nonlinear transformation: $\bk = \pi \circ T$.

Following \cref{app:eq:proof2-aug}, we can write the relationship between the augmented functions as:
\begin{align}
    \gaug \circ \faug \circ \kaug &= \hatgaug \circ \hatfaug, \\
    \underbrace{\hatgaug^{-1} \circ \gaug}_{\h^{-1}_{\texttt{aug}}} \circ \faug \circ \kaug &= \hatfaug, \\
    \h^{-1}_{\texttt{aug}} \circ \faug \circ \kaug &= \hatfaug, \\
\end{align}
where $\h^{-1}_{\texttt{aug}} = [\h^{-1}, \h^{-1}]$. We have shown that both $\h$ and $\bk$ are compositions of permutations and element-wise invertible transformations. Hence, the augmented transition function $\hatfaug$ is equal to the true augmented transition function $\faug$ up to compositions of permutations and element-wise transformations.

\subsection{Alternative Versions of Sufficient Variability Assumption}
\label{app:sec:alternative-assump}

    If the variable $\bu$ is an observed categorical variable (e.g., domain indicator), the assumptions \textit{(A4, A5)} can be written in an alternative form without partial derivatives with respect to $u_l$, similar to \citet{hyvarinen2019nonlinear, yao2021learning}. For example, for the latent states $\z_t$, the alternative version of the \textit{(A4)} takes the form:
    \begin{itemize}
        \item \textbf{Sufficient variability of latent states for a categorical $\bu$ \citep{yao2021learning}.} For any $\z_t$, there exist some $2K+1$ values for $\bu$: $\bu_1, \ldots, \bu_{2K}$, such that the $2K$ vectors $\bv(\z_t, \bu_{j+1}) - \bv(\z_t, \bu_{j})$ with $j=0,1,\ldots, 2K$, are linearly independent where
		\begin{align}
			\bv(\z_t, \bu) = \left( \frac{\partial \eta_{1}(z_{1t}, \bu)}{\partial z_{1t}}, \cdots, \frac{\partial \eta_{K}(z_{Kt}, \bu)}{\partial z_{Kt}}, \frac{\partial^2 \eta_{1}(z_{1t}, \bu)}{\partial z^2_{1t}}, \cdots, \frac{\partial^2 \eta_{K}(z_{Kt}, \bu)}{\partial z^2_{Kt}} \right) \in \R^{2K}.
		\end{align}
    \end{itemize}
    A similar categorical version is provided in \citep[Assumption~3]{hyvarinen2019nonlinear}, while the continuous version is provided in the same work \citep[Appendix~D]{hyvarinen2019nonlinear}.
 
	\section{Variational inference}
	\label{sec:app:vi}
	
	Similar to previous works \citep{yao2021learning, yao2022temporally}, we want to maximize the marginal log-likelihood $\log p(\x_{1:T} | \bu)$ that is obtained by marginalizing over the latent states $\z_{1:T}$ and process noise $\s_{1:T}$:
	\begin{align}
		\log \pth(\x_{1:T} | \bu) = \log \int_{\z, \s} \pth(\x_{1:T}, \z_{0:T}, \s_{1:T} | \bu) \D\z_{0:T} \D\s_{1:T},
	\end{align}
	where we decompose the joint distribution as follows:
	\begin{align}
		\pth(\x_{1:T}, \z_{0:T}, \s_{1:T} | \bu)
		&= \pth(\z_0) \prod_{t=1}^T \pth(\s_t | \bu) \underbrace{\pth(\z_t | \z_{t-1}, \s_{t})}_{\delta \left(\z_t - \f(\s_t,\z_{t-1}) \right)} \pth(\x_t | \z_t).
	\end{align}
	Note that the state transitions $\pth(\z_t | \z_{t-1}, \s_{t})$ are assumed to be deterministic. The above integral is intractable due to non-linear dynamics $\f$ and observation $\g$ functions. As typically done with the deep latent variable models, we approximate the log marginal likelihood by a variational lower bound, i.e., we introduce an amortized approximate posterior distribution $\qph(\z_{0:T}, \s_{1:T} | \x_{1:T}, \bu)$ that decomposes as follows:
	\begin{align}
		\qph(\z_{0:T}, \s_{1:T} | \x_{1:T}, \bu) = \qph(\z_0 | \x_{1:T}, \bu) \prod_{t=1}^T \underbrace{\qph(\s_t | \z_{0:t-1}, \s_{1:t-1}, \x_{1:T}, \bu)}_{\qph(\s_t | \z_{t-1}, \x_{1:t})} \underbrace{\qph(\z_t | \z_{0:t-1}, \s_{1:t}, \bu)}_{\qph(\z_t | \z_{t-1}, \s_{t})}
	\end{align}
	We simplify the variational posterior $q(\s_t | \cdot)$ as $\qph(\s_t | \z_{0:t-1}, \s_{1:t-1}, \x_{1:T}, \bu) = \qph(\s_t | \z_{t-1}, \x_{1:t})$, corresponding to a filtering distribution. As in the generative model, we choose $\qph(\z_t | \z_{t-1}, \s_{t}) = \pth(\z_t | \z_{t-1}, \s_{t}) = \delta \left(\z_t - \f(\s_t,\z_{t-1}) \right)$:
	\begin{align}
		\qph(\z_{0:T}, \s_{1:T} | \x_{1:T}, \bu) = \qph(\z_0 | \x_{1:T}) \prod_{t=1}^T \qph(\s_t | \z_{t-1}, \x_{1:t}) \pth(\z_t | \z_{t-1}, \s_t),
	\end{align}
	where the functional forms of the densities $\qph(\z_0|\cdot)$ and $\qph(\s_t|\cdot)$ are chosen as diagonal Gaussian distributions whose parameters are computed by recurrent neural networks. The variational lower bound takes the following form:
	\begin{align}
		\mathcal{L}(\theta, \phi) &= \EX_{\qph(\z_{0:T},\s_{1:T})}\left[\log \pth(\x_{1:T} | \z_{1:T}) + \log \frac{\pth(\z_{0:T}, \s_{1:T} | \bu)}{\qph(\z_{0:T}, \s_{1:T} | \x_{1:T}, \bu)} \right] \\
		&= \underbrace{\sum_{t=1}^T \EX_{\qph(\z_{t})}[\log \pth(\x_{t} | \z_{t})]}_{\text{Reconstruction term, } \mathcal{L}_{R}} + \underbrace{\EX_{\qph(\z_{0:T},\s_{1:T})}\left[\log \frac{\pth(\z_{0:T}, \s_{1:T} | \bu)}{\qph(\z_{0:T}, \s_{1:T} | \x_{1:T}, \bu)} \right]}_{\text{KL term, } \mathcal{L}_{KL}}.
	\end{align}
	The reconstruction term can easily be computed in a variational auto-encoder framework. Below, we provide the derivation of of the KL term:
\begin{align}
    \mathcal{L}_{KL} &= \EX_{\qph(\z_{0:T},\s_{1:T})}\left[ \log \frac{\pth(\z_{0})}{\qph(\z_{0} | \x_{1:T}, \bu)} + \log \frac{\pth(\z_{1:T}, \s_{1:T} | \bu)}{\qph(\z_{1:T}, \s_{1:T} | \x_{1:T}, \bu)} \right] \\
    &= - D_{KL}(\qph(\z_{0} | \x_{1:T}, \bu) || \pth(\z_{0})) + \EX_{\qph(\z_{0:T},\s_{1:T})}\left[ \log \frac{\pth(\z_{1:T}, \s_{1:T} | \bu)}{\qph(\z_{1:T}, \s_{1:T} | \x_{1:T}, \bu)} \right] \\
    &= - D_{KL}(\qph(\z_{0} | \x_{1:T}, \bu) || \pth(\z_{0})) + 
    \EX_{\qph(\z_{0:T},\s_{1:T})}\left[ 
    \sum_{t=1}^T \log 
    \frac{\pth(\s_t | \bu) \pth(\z_t | \z_{t-1}, \s_{t})}
    {\qph(\s_{t} | \z_{t-1}, \x_{1:T}, \bu) \qph(\z_t | \z_{t-1}, \s_t)} 
    \right] \\
    &= - D_{KL}(\qph(\z_{0} | \x_{1:T}, \bu) || \pth(\z_{0})) +  \sum_{t=1}^T \EX_{\qph(\z_{0:T},\s_{1:T})}\left[\log \frac{\pth(\s_t | \bu)}
    {\qph(\s_{t} | \z_{t-1}, \x_{1:T}, \bu)} 
    \right] \\
    &= - D_{KL}(\qph(\z_{0} | \x_{1:T}, \bu) || \pth(\z_{0})) +  \sum_{t=1}^T \underbrace{\EX_{\qph(\s_{t},\z_{t-1} | \z_{t-2}, \s_{t-1}, \x_{1:T}, \bu)} \left[\log \frac{\pth(\s_{t} | \bu) }
    {\qph(\s_{t} | \z_{t-1}, \x_{1:T}, \bu)} \right]}_{-\EX_{\qph(\z_{t-1} | \z_{t-2}, \s_{t-1})}[D_{KL}(\qph(\s_{t} | \z_{t-1}, \x_{1:T}, \bu) || \pth(\s_{t} | \bu))]}
\end{align}

\begin{figure}
    \centering
    \includegraphics[width=\linewidth]{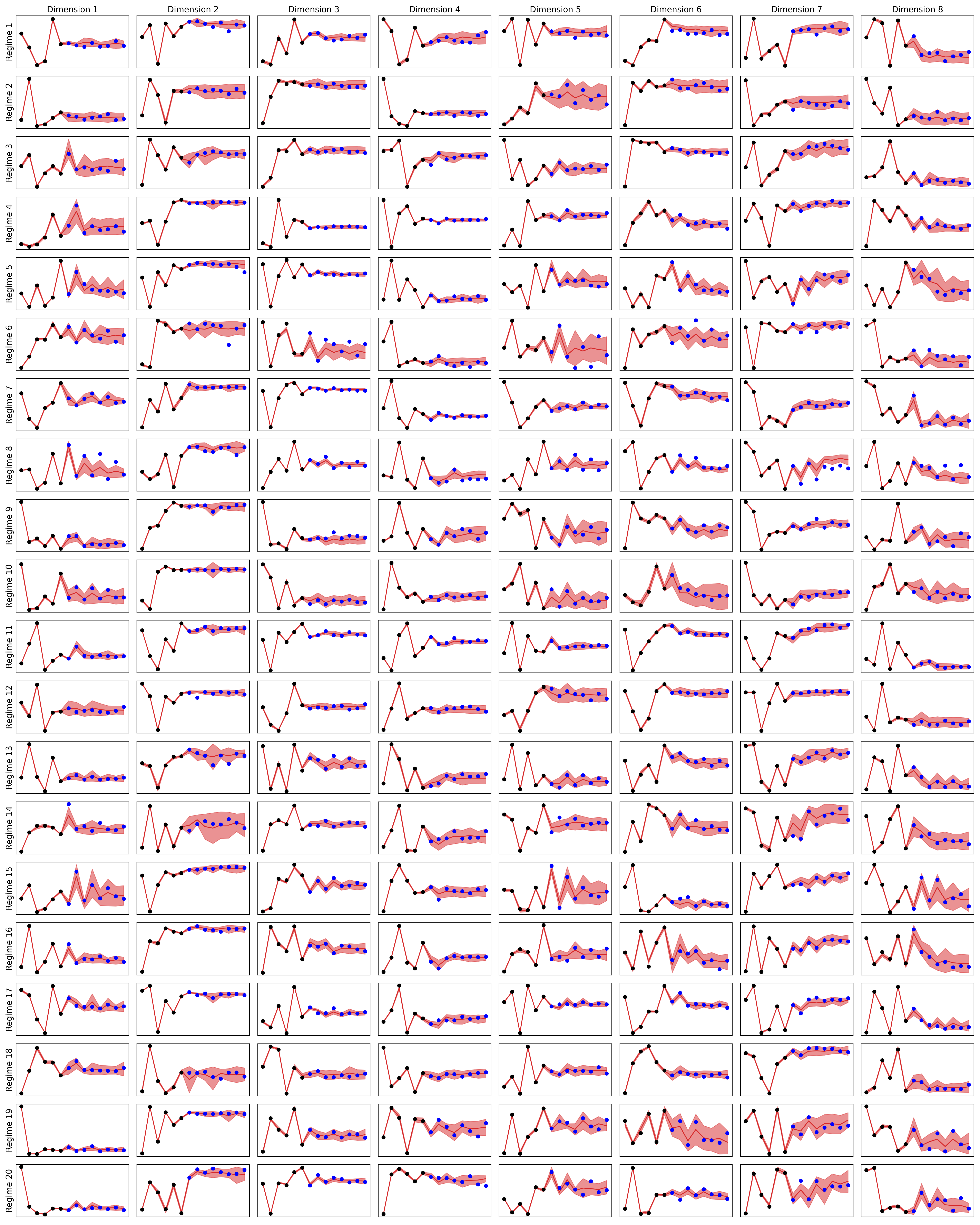}
    \caption{Extended version of Figure~\ref{fig:uncertainty}}
    \label{fig:app:uncertainty}
\end{figure}

\begin{figure}
    \centering
    \includegraphics[width=\linewidth]{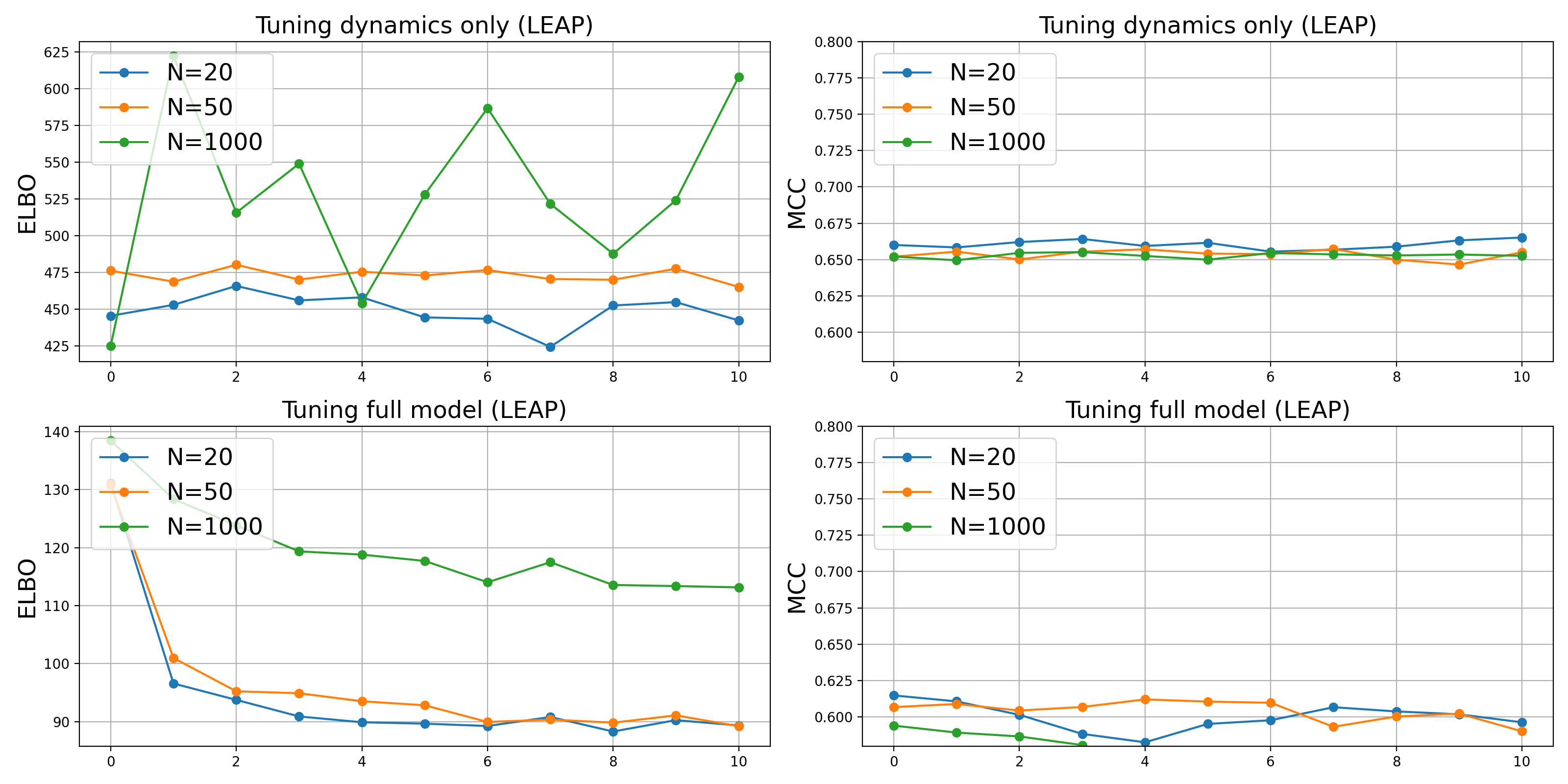} \\
    \includegraphics[width=\linewidth]{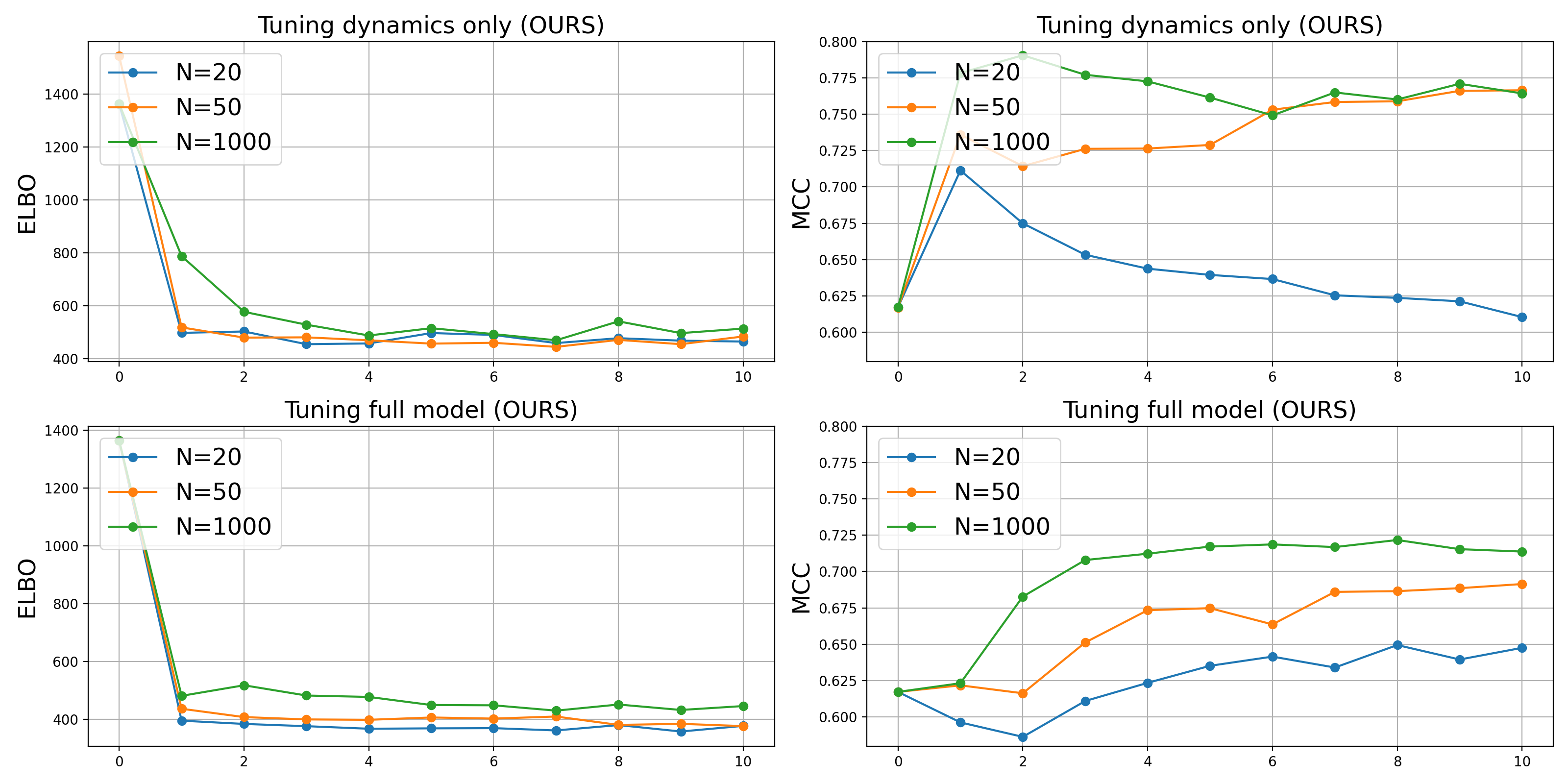}
    \caption{We trace the $\elbo$ and $\mcc$ metrics while adapting LEAP and our approach to small Cartpole adaptation datasets. $x$-axis shows the gradient steps ($\times 1000$) }
    \label{fig:app:adaptation}
\end{figure}
 
	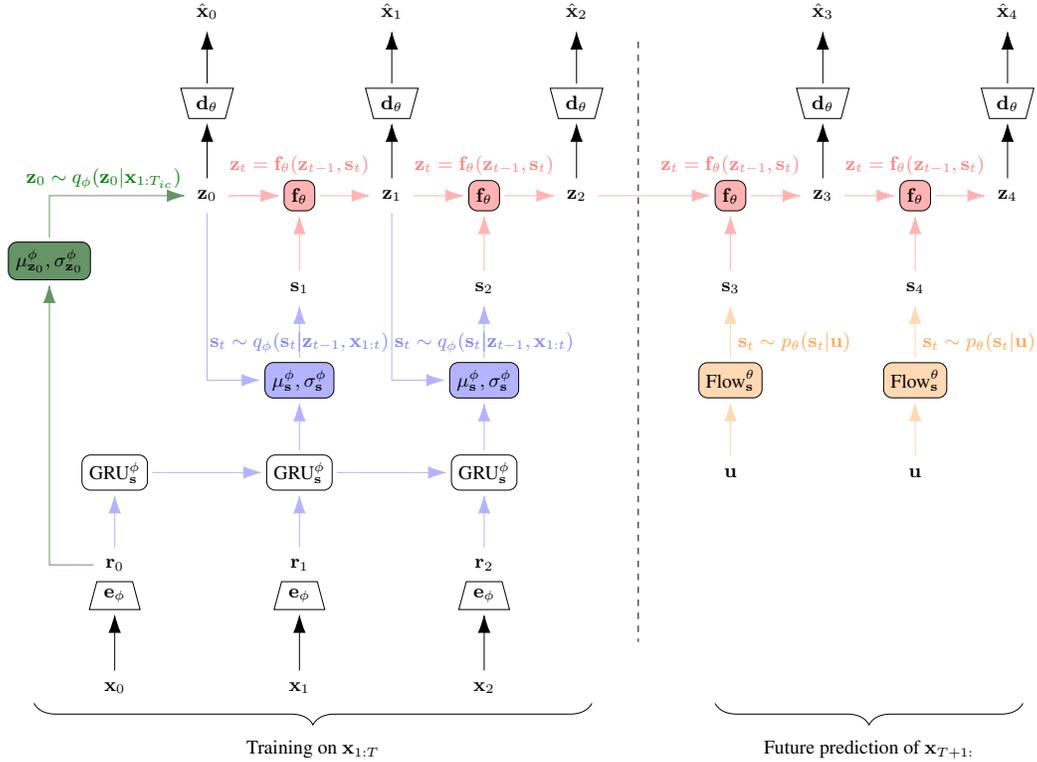
\begin{figure}
		\centering
        \resizebox{\textwidth}{!}{
            \begin{tikzpicture}[every node/.style={font=\small}, node distance=1.5cm and 1.5cm]
\colorlet{transition}{red!30!white}
\colorlet{transitiontext}{red!50!white}
\colorlet{noise}{orange!30}
\colorlet{noisetext}{orange!70}
\colorlet{encoder}{white}
\colorlet{decoder}{white}
\colorlet{muinitial}{green!30!black!60!white}
\colorlet{muinitialtext}{green!50!black}
\colorlet{gru}{white}
\colorlet{encline}{blue!30}
\colorlet{munoise}{blue!30}
\colorlet{munoisetext}{blue!50}

    % Styles
    \tikzstyle{trapezoid} = [trapezium, draw, text centered, minimum height=1em]
    \tikzstyle{encoder} = [trapezoid, trapezium angle=75, fill=encoder]
    \tikzstyle{decoder} = [trapezoid, trapezium angle=105, fill=decoder]
    \tikzstyle{block} = [rectangle, draw, rounded corners, text centered, minimum height=1em]
    \tikzstyle{gru} = [block, fill=gru]
    \tikzstyle{transition} = [block, fill=transition]
    \tikzstyle{noise} = [block, fill=noise]

    \tikzstyle{line} = [draw, -{Latex[length=3mm, width=2mm]}, shorten >=2pt, shorten <=2pt]
    \tikzstyle{fline} = [line, color=transition]
    \tikzstyle{eline} = [line, color=encline]
    \tikzstyle{sline} = [line, color=noise]
    \tikzstyle{muinitialline} = [line, color=muinitial]

    \tikzstyle{bracebelow} = [decorate,decoration={brace,amplitude=10pt,mirror}]
    \tikzstyle{bracelabel} = [midway,yshift=-0.75cm]

    % Nodes
    \node[] (z0) {$\vz_0$};
    \node[right of=z0, transition] (f0) {$\vf_\theta$};
    \node[right of=f0] (z1) {$\vz_1$};
    \node[right of=z1, transition] (f1) {$\vf_\theta$};
    \node[right of=f1] (z2) {$\vz_2$};
    \node[right of=z2, transition, xshift=1cm] (f2) {$\vf_\theta$};
    \node[right of=f2] (z3) {$\vz_3$};
    \node[right of=z3, transition] (f3) {$\vf_\theta$};
    \node[right of=f3] (z4) {$\vz_4$};
    
    \path [fline] (z0) -- (f0);
    \foreach \i in {0, ..., 3} {
        \node[above of=f\i, yshift=-1cm, color=transitiontext] (f\i{}label) {$\vz_t = \vf_\theta(\vz_{t-1}, \vs_t)$};
    
        \pgfmathtruncatemacro{\nexti}{\i + 1}
        \node[below of=f\i] (s\nexti) {$\vs_\nexti$};
        \path [fline] (s\nexti) -- (f\i);
        \path [fline] (z\i) -- (f\i);
        \path [fline] (f\i) -- (z\nexti);
    }

    \foreach \i in {0, ..., 4} {
        \node[above of=z\i, decoder] (d\i) {$\vd_\theta$};
        \node[above of=d\i] (xhat\i) {$\hat{\vx}_\i$};
        \path [line] (z\i) -- (d\i);
        \path [line] (d\i) -- (xhat\i);
        }

    \node[block, below left of=z0, xshift=-1.5cm, fill=muinitial] (mu0) {$\mu_{\vz_0}^\phi, \sigma_{\vz_0}^\phi$};

    \foreach \i in {1, 2} {
        \node[block, below of=s\i, fill=munoise] (mu\i) {$\mu_{\vs}^\phi, \sigma_{\vs}^\phi$};
        \node[gru, below of=mu\i] (gru\i) {GRU$_{\vs}^\phi$};
    }
    \node[gru, left of=gru1, xshift=-1.5cm] (gru0) {GRU$_{\vs}^\phi$};

    \foreach \i in {0, ..., 2} {
        \node[below of=gru\i] (r\i) {$\vr_\i$};
        \node[encoder, below of=r\i, yshift=1cm] (e\i) {$\ve_\phi$};
        \node[below of=e\i] (x\i) {$\vx_\i$};
        \path[line] (x\i) -- (e\i);
        \path[eline] (r\i) -- (gru\i);
    }

    % \foreach \i in {1, 2} {
    %     \node[left of=s\i, noise, xshift=-1cm] (flow\i) {Flow${}_\vs^\theta$};
    %     \path[sline] (flow\i) -- node[above, xshift=0.3cm, color=noisetext] {$p_\theta(\vs_t | \vu)$} (s\i);    
    % }
    \foreach \i in {3, 4} {
        \node[below of=s\i, noise] (flow\i) {Flow${}_\vs^\theta$};
        \path[sline] (flow\i) -- node[right, yshift=-5pt, color=noisetext] {$\vs_t \sim p_\theta(\vs_t | \vu)$} (s\i);    
    }

    \foreach \i in {3, ..., 4} {
        \node[below of=flow\i] (u\i) {$\vu$};
        \path[sline] (u\i) -- (flow\i);
    }
    
    \path [muinitialline] (r0) -| (mu0);
    % \path [muinitialline] ($(r1) + (-0.3,0.1)$) -| (mu0);
    \path [muinitialline] (mu0) |- node[above right, xshift=-0.5cm, color=muinitialtext] {\small $\vz_0 \sim q_\phi(\vz_0 | \vx_{1:T_{ic}})$} (z0);

    \foreach \i in {0, 1} {
        \pgfmathtruncatemacro{\nexti}{\i + 1}
        \path[eline] (gru\i) -- (gru\nexti);
        \path[eline] (gru\nexti) -- (mu\nexti);
        \path[eline] (mu\nexti) -- node[midway, yshift=-5pt, color=munoisetext, fill=none] {$\vs_t \sim q_\phi(\vs_t | \vz_{t-1}, \vx_{1:t})$} (s\nexti);
        \path[eline] (z\i) |- (mu\nexti);
    }
    
    % Dashed line
    \draw[dashed] ($(d2.north east) + (0.75, 0.75)$) -- ($(d2.south east) + (0.75, -8.5)$);
    
    % Braces
    \coordinate (brace1left) at (mu0 |- x0);
    \coordinate (brace1right) at (z2 |- x1);
    \draw [bracebelow] ($(brace1left) + (-0.25,-0.25)$) -- ($(brace1right) + (0.25,-0.25)$) node[bracelabel] {Training on $\vx_{1:T}$};
    \coordinate (brace2left) at (f2 |- x0);
    \coordinate (brace2right) at (z4 |- x1);
    \draw [bracebelow] ($(brace2left) + (-0.25,-0.25)$) -- ($(brace2right) + (0.25,-0.25)$) node[midway,yshift=-0.75cm] {Future prediction of $\vx_{T+1:}$};
\end{tikzpicture}
        }
		\caption{Diagram of the model architecture: In training, the observation $\x$ is passed through the encoder $\mathbf{e}_\phi$ to get the representation $\mathbf{r}$. We learn the distribution over the initial latent state $\z_0$ conditional on the representations of the first $T_{ic}$ observations (in the diagram, we show $T_{ic}=1$; in our experiments, we use $T_\text{ic}=2$). The latent state is decoded by the decoder $\mathbf{d}_\theta$ to produce the predicted observation $\hat{\x}$ (which is trained to match the corresponding actual observation). The next value of the latent state is computed by the transition function $\f_\theta$, which depends both on the previous state and on the process noise $\s$. In training, the process noise $\s$ is sampled from the variational posterior that depends on the previous state as well as on the representation created by a recurrent neural network (GRU) that has received up to the current observation. In future prediction, the process noise $\s$ is sampled from the prior, which is a learned normalizing flow.
        }
		\label{fig:app:arch}
	\end{figure}

\section{Architecture and optimization details}
\label{sec:app:arc}
We optimize our model with Adam optimizer. We chose all hyperparameters for our method, two versions of LEAP and KalmanVAE with cross-validation. In particular, we performed random search as well as Bayesian optimization over learning rate, weight regularization, the number of layers in all MLPs, and latent dimensionality.

\subsection{Synthetic data experiments}
\begin{itemize}
    \item \textbf{Encoder}: Below, the output of encoder\_base\_layer goes into encoder\_rnn\_layer and s\_encoder.
    \begin{itemize}
        \item (encoder\_base\_layer (MLP)): 
        \begin{itemize}
            \item \texttt{Linear(in\_features=8, out\_features=64, bias=True)}
            \item \texttt{LeakyReLU(negative\_slope=0.2)}
\item \texttt{Linear(in\_features=64, out\_features=64, bias=True)}
            \item \texttt{LeakyReLU(negative\_slope=0.2)}
\item \texttt{Linear(in\_features=64, out\_features=64, bias=True)}
            \item \texttt{LeakyReLU(negative\_slope=0.2)}
            \item \texttt{Linear(in\_features=64, out\_features=64, bias=True)}
        \end{itemize}
        \item   (encoder\_rnn\_layer): \texttt{GRU(in=64, hidden\_dim=64, output\_size=64)}
        \item  (ic\_encoder): 
        \begin{itemize}
            \item \texttt{Linear(in\_features=256, out\_features=64, bias=True)}
            \item \texttt{LeakyReLU(negative\_slope=0.2)}
            \item \texttt{Linear(in\_features=64, out\_features=64, bias=True)}
            \item \texttt{LeakyReLU(negative\_slope=0.2)}
            \item \texttt{Linear(in\_features=64, out\_features=16, bias=True)}
        \end{itemize}
      \item (s\_encoder (MLP)): 
      \begin{itemize}
          \item \texttt{Linear(in\_features=80, out\_features=64, bias=True)}
        \item \texttt{LeakyReLU(negative\_slope=0.2)}, 
          \item \texttt{Linear(in\_features=64, out\_features=64, bias=True)}
          \item \texttt{LeakyReLU(negative\_slope=0.2)}, 
          \item \texttt{Linear(in\_features=64, out\_features=16, bias=True)}
      \end{itemize}
    \end{itemize}
    
    \item \textbf{Decoder (MLP)}
    \begin{itemize}
        \item \texttt{Linear(in\_features=8, out\_features=64, bias=True)}
        \item \texttt{LeakyReLU(negative\_slope=0.2)}, 
        \item \texttt{Linear(in\_features=64, out\_features=64, bias=True)}
        \item \texttt{LeakyReLU(negative\_slope=0.2)}
        \item \texttt{Linear(in\_features=64, out\_features=8, bias=True))}
    \end{itemize}

    \item \textbf{Transition function}: Here, we consider 8 different MLPs, each of which has the following architecture:
    \begin{itemize}
        \item \texttt{Linear(in\_features=17, out\_features=64, bias=True)}
        \item \texttt{LeakyReLU(negative\_slope=0.2)}, 
        \item \texttt{Linear(in\_features=64, out\_features=64, bias=True)}
        \item \texttt{LeakyReLU(negative\_slope=0.2)}
        \item \texttt{Linear(in\_features=64, out\_features=1, bias=True))}
    \end{itemize}
\end{itemize}

\subsection{Cartpole experiments}
\begin{itemize}
    \item \textbf{Encoder}
    \begin{itemize}
        \item \texttt{Conv2d(in\_channels=3, num\_filter=32, kernel=3, stride=2, pad=1)}
        \item \texttt{GeLU()}
        \item \texttt{Conv2d(in\_channels=32, num\_filter=32, kernel=3, stride=2, pad=1)}
        \item \texttt{GeLU()}
        \item \texttt{Conv2d(in\_channels=32, num\_filter=64, kernel=3, stride=2, pad=1)}
        \item \texttt{GeLU()}
        \item \texttt{Conv2d(in\_channels=64, num\_filter=64, kernel=3, stride=2, pad=1)}
        \item \texttt{GeLU()}
        \item \texttt{Flatten()}
        \item \texttt{Linear(1024,8)}
    \end{itemize}
    \item \textbf{Decoder}
    \begin{itemize}
        \item \texttt{Linear(8,1024)}
        \item \texttt{UnFlatten(4x4x64)}
        \item \texttt{ConvTranspose2d(in\_channels=64, num\_filter=64, kernel=3, stride=2, pad=1, output\_padding=1)}
        \item \texttt{GeLU()}
        \item \texttt{ConvTranspose2d(in\_channels=64, num\_filter=32, kernel=3, stride=2, pad=1, output\_padding=1)}
        \item \texttt{GeLU()}
        \item \texttt{ConvTranspose2d(in\_channels=32, num\_filter=32, kernel=3, stride=2, pad=1, output\_padding=1)}
        \item \texttt{GeLU()}
        \item \texttt{ConvTranspose2d(in\_channels=32, num\_filter=1, kernel=3, stride=2, pad=1, output\_padding=1)}
        \item \texttt{sigmoid()}
    \end{itemize}
    
    \item \textbf{Transition function}: Here, we consider 8 different MLPs, each of which has the following architecture:
    \begin{itemize}
        \item \texttt{Linear(in\_features=17, out\_features=64, bias=True)}
        \item \texttt{LeakyReLU(negative\_slope=0.2)}
        \item \texttt{Linear(in\_features=64, out\_features=64, bias=True)}
        \item \texttt{LeakyReLU(negative\_slope=0.2)}
        \item \texttt{Linear(in\_features=64, out\_features=1, bias=True))}
    \end{itemize}
    \item \textbf{Normalizing Flow:} As the nonstationary prior for the noise variables, we use $1D$ conditional normalizing flows, which are 1-layer neural spline flows conditioned on the auxiliary variable $\bu$. Before taken as, the auxiliary variable $\bu$ is embedded. This is done by a single linear layer with 32 dimensions in the synthetic experiments, and an MLP with the following architecture in the cartpole experiment:
    \begin{itemize}
        \item \texttt{Linear(in\_features=7, out\_features=64, bias=True)}
        \item \texttt{LeakyReLU(negative\_slope=0.2)}
        \item \texttt{Linear(in\_features=64, out\_features=32, bias=True)}
    \end{itemize}
\end{itemize}

\end{document}